\documentclass[10pt,twocolumn,letterpaper]{article}

\usepackage[pagenumbers]{iccv} 

\definecolor{iccvblue}{rgb}{0.21,0.49,0.74}
\usepackage[pagebackref,breaklinks,colorlinks,allcolors=iccvblue]{hyperref}
\usepackage{url}
\usepackage{booktabs}
\usepackage{xspace}
\usepackage{xcolor}
\usepackage{graphicx}
\usepackage{caption}
\usepackage{subcaption}
\usepackage{enumitem}
\usepackage{wrapfig}
\usepackage{makecell}
\usepackage{multirow}

\newcommand{\hide}[1]{\phantom{}} 

\newcommand{\ali}[1]{\phantom{}} 
\makeatletter

\definecolor{demphcolorinline}{gray}{.3}

\definecolor{demphcolor1}{gray}{.6}

\def\paperID{9110} 
\def\confName{ICCV}
\def\confYear{2025}

\title{Stable Diffusion Models are Secretly Good at Visual In-Context Learning}

\newcommand{\authorskip}{\hspace{8mm}}

\author{Trevine Oorloff\textsuperscript{1,2\thanks{This work was completed during internship at Apple}}
\authorskip Vishwanath Sindagi\textsuperscript{1}
\authorskip Wele Gedara Chaminda Bandara\textsuperscript{1} \\
\authorskip Ali Shafahi\textsuperscript{1}
\authorskip Amin Ghiasi\textsuperscript{1}
\authorskip Charan Prakash\textsuperscript{1}
\authorskip Reza Ardekani\textsuperscript{1}
\vspace{3mm}\\ 
\textsuperscript{1}Apple \hspace{8mm} \textsuperscript{2}University of Maryland - College Park\\
}

\begin{document}
\maketitle
\begin{abstract}
Large language models (LLM) in natural language processing (NLP) have demonstrated great potential for in-context learning (ICL) --- the ability to leverage a few sets of example prompts to adapt to various tasks without having to explicitly update the model weights. 
ICL has recently been explored for computer vision tasks with promising early outcomes. These approaches involve specialized training and/or additional data that complicate the process and limit its generalizability. In this work, we show that off-the-shelf Stable Diffusion models can be repurposed for visual in-context learning (V-ICL). Specifically, we formulate an in-place attention re-computation within the self-attention layers of the Stable Diffusion architecture that explicitly incorporates context between the query and example prompts. Without any additional fine-tuning, we show that this repurposed Stable Diffusion model is able to adapt to six different tasks: foreground segmentation, single object detection, semantic segmentation, keypoint detection,  edge detection, and colorization. 
For example, the proposed approach improves the mean intersection over union (mIoU) for the foreground segmentation task on Pascal-5i dataset by 8.9\% and 3.2\% over recent methods such as Visual Prompting and IMProv, respectively. Additionally, we show that the proposed method is able to effectively leverage multiple prompts through ensembling to infer the task better and further improve the performance.  
\end{abstract}    
\section{Introduction}
\label{sec:intro}

\begin{figure}[tbp]
    \centering
    \includegraphics[width=0.95\linewidth]{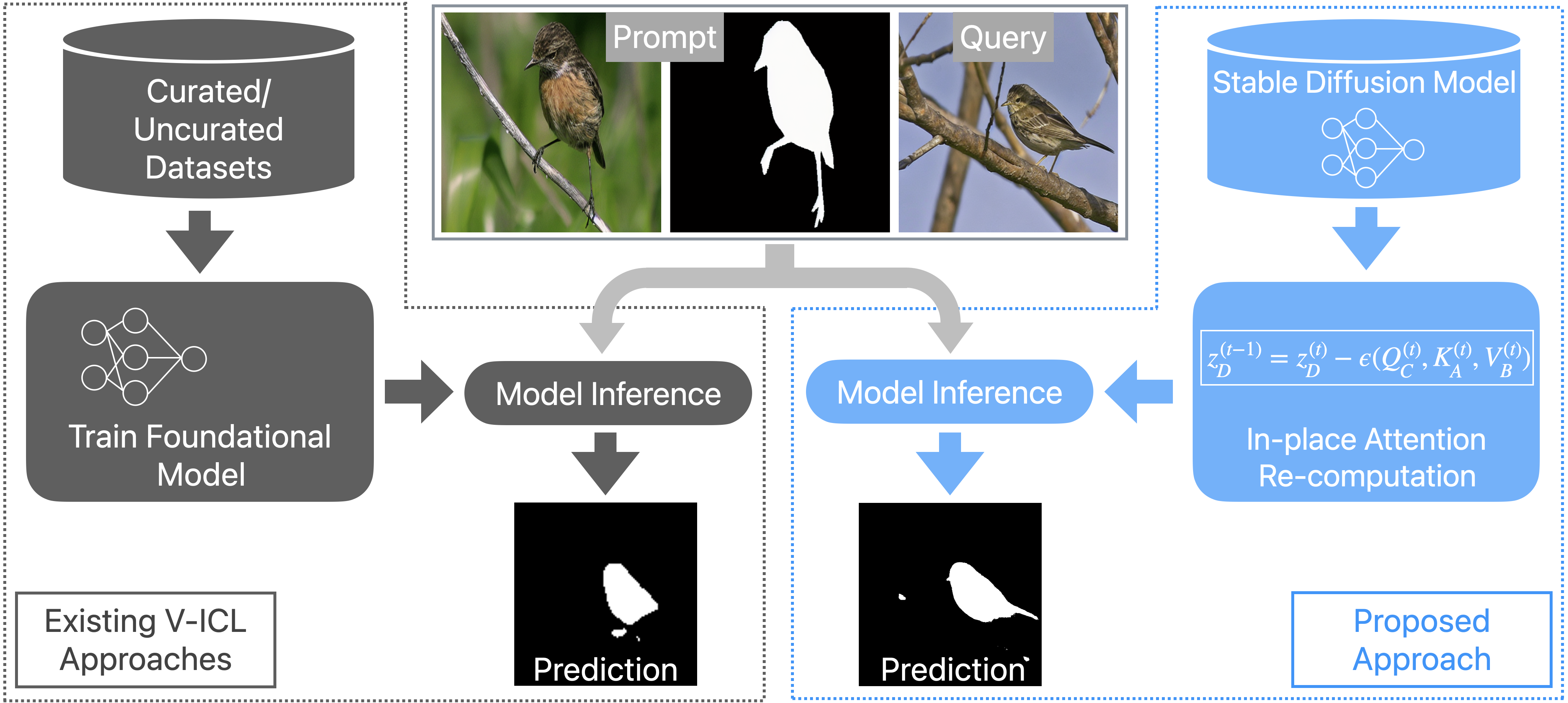}
   \caption{Existing V-ICL methods require specialized training of foundation models on curated/uncurated data before using them for novel downstream tasks. In contrast, our approach repurposes an off-the-shelf Stable Diffusion model without any additional training or data.
   }\vspace{-15pt}
    \label{fig:teaser}
\end{figure}

\textit{In-context learning (ICL)} refers to the paradigm where a foundation model leverages exemplar source-target pair(s), known as \textit{prompts}, to infer the task and perform the inferred task on an input, known as \textit{query}. 
ICL is an emergent property of large language models (LLMs) and has been widely explored in the natural language processing (NLP) domain \citep{wei2022emergent, brown2020gpt3, hao2022language, touvron2023llama}. 
ICL facilitates a model to adapt to novel or out-of-domain tasks without the need for fine-tuning which not only eliminates task-specific training but also reduces dependency on task-specific annotated datasets. 
Recent research, such as \cite{bar2022visualprompting, xu2023improv, wang2023painter, wang2023seggpt, wang2023promptdiff, liu2023explicit,wang2024skeleton}, have made promising attempts to harness the potential of in-context learning for computer vision tasks. These works can be broadly divided into two categories based on the datasets that are used for training: (1)  uncurated datasets (\eg CVF \citep{bar2022visualprompting}, S2CV \citep{xu2023improv}) and (2) curated/annotated task-related datasets (\eg COCO \citep{lin2014coco}, NYUDv2 \citep{silberman2012nyudv2}).

\begin{figure}[tbp]
    \centering
    \begin{subfigure}{0.7\linewidth}
        \centering
        \includegraphics[width=\linewidth]{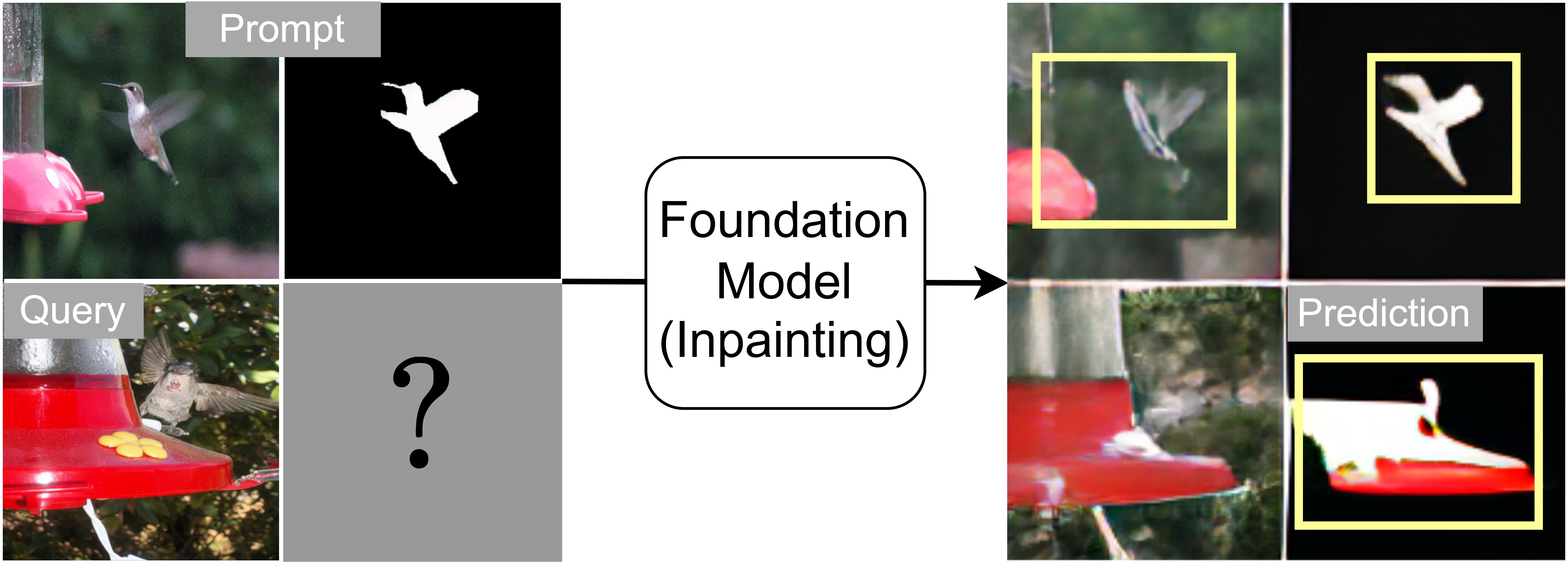}
        \caption{}
        \label{fig:vp_drawbacks}
    \end{subfigure}
    \par
    \centering
    \scalebox{0.7}{
    \begin{minipage}{\linewidth}
    \begin{subfigure}{0.4\linewidth}
        \centering
        \includegraphics[width=\linewidth]{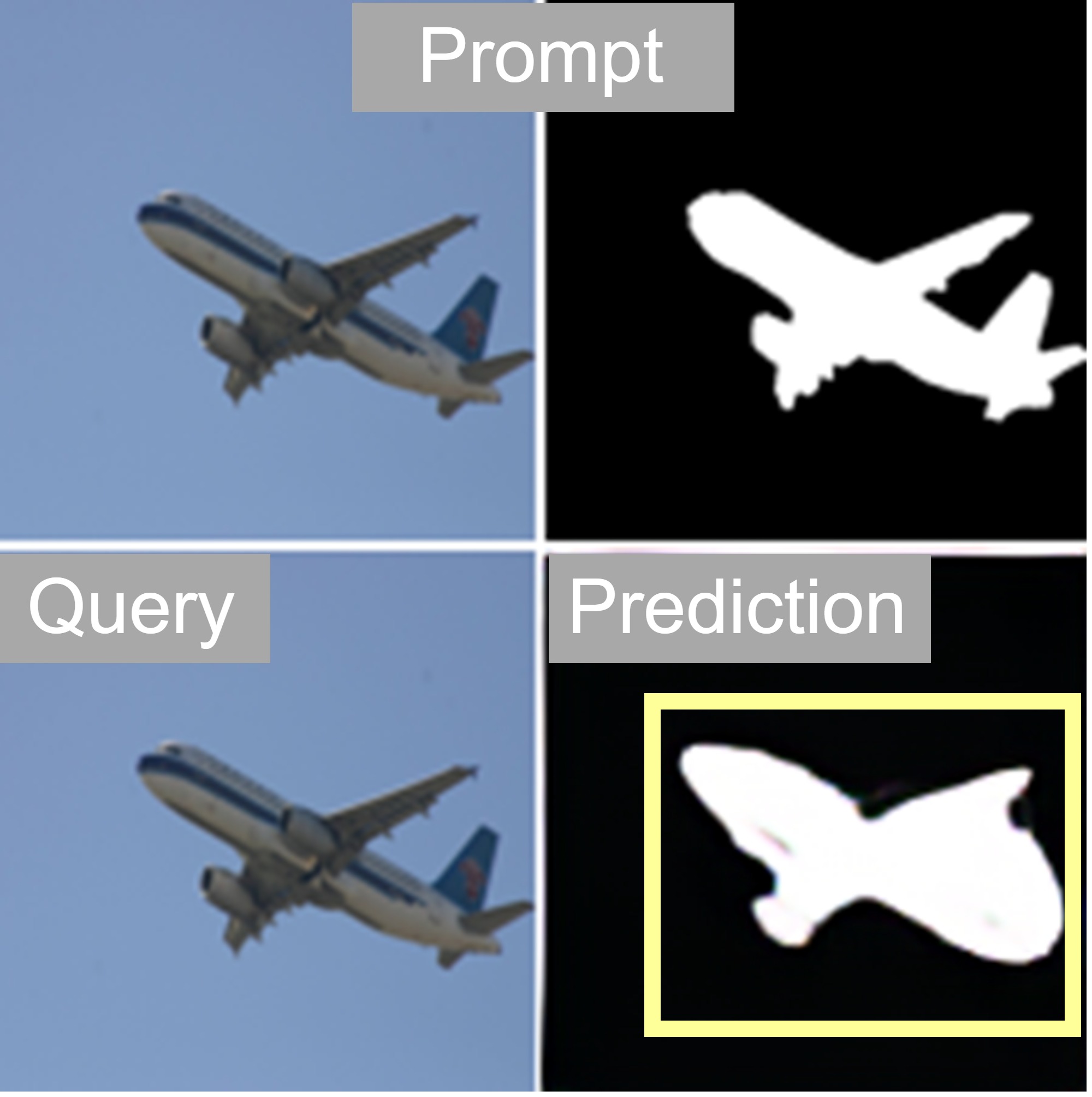}
        \caption{}
        \label{fig:vp_drawbacks_toy}
    \end{subfigure}
    \hfill
    \begin{subfigure}{0.4\linewidth}
        \centering
        \includegraphics[width=\linewidth]{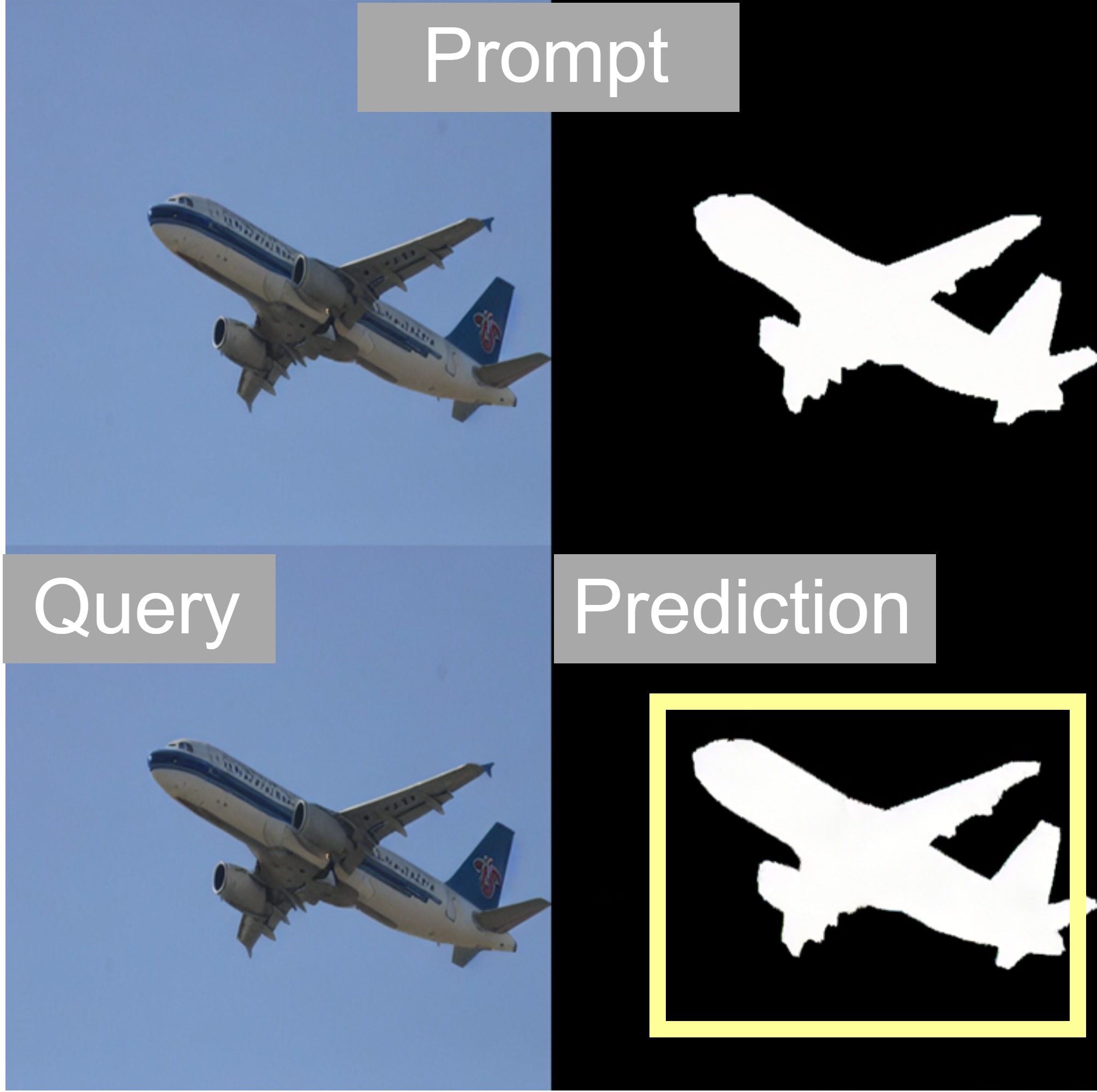}
        \caption{}
        \label{fig:vp_drawbacks_toy_ours}
    \end{subfigure}
    \end{minipage}
    }
    \caption{(a) Existing approaches like Visual Prompting \citep{bar2022visualprompting} train a foundation model with an inpainting objective on uncurated data. At inference, a grid of query and example prompt are input to the model. (b) Visual Prompting \citep{bar2022visualprompting} struggles to predict accurately even when the prompt is the same as the query. (c) The proposed approach, when presented with the same prompt as the query, is able to fully leverage the prompt to make accurate predictions.
    }\vspace{-12pt}
    \label{fig:drawbacks}
\end{figure}

The first category of approaches such as Visual Prompting \citep{bar2022visualprompting} and Improv \citep{xu2023improv} enable foundation models to perform in-context learning by training with an inpainting loss on uncurated datasets  (\eg CVF, S2CV).
These datasets consist of images extracted from computer vision papers and consist of grid-like structures of source-target pairs. 
The combination of this structured input and random masking during training allows the model to implicitly learn the relationships between the source and target images. 
During inference, when the model is presented with a grid-like canvas of the query image and the example prompt(s) (source-target pairs) along with the location of the output for the query image masked-out (see \cref{fig:vp_drawbacks}), the model is expected to inpaint the prediction. 

The second category of approaches \citep{wang2023painter, wang2023seggpt, wang2023promptdiff, liu2023explicit,wang2024skeleton} enable a foundation model for in-context learning by allowing the model to train on curated datasets (\eg COCO, NYUDv2) related to the out-of-domain tasks that the model is expected to adapt to. While these methods do not employ task-specific loss functions, they still rely on task-representative datasets. For example, to enable the Painter \citep{wang2023painter} model to adapt to open-vocabulary segmentation (\eg FSS-1000 \citep{li2020fss}), the model is trained on a related dataset like the COCO segmentation dataset. In other words, as long as the model has encountered a task during training, it can perform that task on query images from unseen categories. However, this implies that the model needs to access large annotated datasets during training thereby undermining the key advantage of V-ICL.
Furthermore, the generalization of such models is limited to tasks related to those seen during training.  These methods\footnote{Refer Sec. B of supp. for a thorough discussion on related work.} show promising early outcomes of vision foundation models on unseen tasks, however, achieving similar success as  LLMs in NLP domain with off-the-shelf vision foundation models (\eg Stable Diffusion \citep{rombach2022stablediff}, ViT \citep{dosovitskiy2020vit}) remains an area yet to be fully explored.

Although these existing V-ICL approaches show promising outcomes, they involve  the use of additional training steps and/or data from out-of-domain tasks to which the model is being used to adapt. 
Different from existing approaches, in this work, we focus on an approach that is more in-line with the ICL  methods from the NLP community. 
Specifically, we address the question --- \textit{Can we use off-the-shelf foundation models for visual in-context learning without any additional training steps or data?} 
To this end, we propose the Stable Diffusion based visual in-context learning (SD-VICL) pipeline where we show that an off-the-shelf Stable Diffusion model can be repurposed, without any fine-tuning, to adapt to several new out-of-domain tasks (see \cref{fig:teaser}). 
To achieve this, we formulate an in-place attention re-computation within the self-attention layers of the Stable Diffusion architecture that explicitly incorporates the context between the query and example prompts. 
This formulation is based on the drawbacks (\cref{fig:drawbacks}) of existing approaches like Visual Prompting \citep{bar2022visualprompting} where these models are unable to fully leverage the example prompt. 

Furthermore, we note that existing approaches such as \citet{bar2022visualprompting} and \citet{xu2023improv} demonstrate the benefits of using multiple example prompts to boost the prediction abilities of the V-ICL foundation models. 
However, their approach of ensembling in the image space by creating a composite of multiple examples comes at the cost of reduced the resolution per prompt --- limiting the potential of prompt ensembling. 
To address this issue, inspired by feature ensembling in SegGPT \citep{wang2023seggpt}, we perform prompt ensembling in the latent space for in-context learning. 
Note that the feature ensembling in SegGPT involves averaging of query features after each attention layer which results in uniform weighting of all prompts. In contrast, we perform implicitly-weighted prompt ensembling within each attention layer thereby enabling the model to better infer the information from multiple prompts.

We conduct extensive evaluations to demonstrate the effectiveness of the proposed approach.
Specifically, we show that a pre-trained Stable Diffusion model has the ability to adapt to various tasks like foreground segmentation, single object detection, semantic segmentation, keypoint detection, edge detection, and colorization. 

To summarize, the main contributions of this work are as follows,
\begin{itemize}[leftmargin=*]
    \item The first training-free method to enable visual in-context learning properties within a foundation model, setting a new direction for visual in-context learning research.
    \item A novel pipeline that explicitly incorporates the context between the query image and prompts by introducing an in-place attention re-computation within the self-attention layers of an off-the-shelf Stable Diffusion model. 
    \item Extensive evaluations of the proposed approach to demonstrate its ability to generalize to multiple out-of-domain tasks. 
    \item Implicitly-weighted prompt ensembling, which enables effective use of multiple prompts for V-ICL.
\end{itemize}
\section{Method} \label{sec:method}

\subsection{Motivation and Preliminaries}\label{sec:background}

In ICL, a foundation model is expected to infer the {task} based on a few source-target example prompts and predict the appropriate output for the query image.   
More importantly, as demonstrated in the NLP community, in-context learning is supposed to be an emergent property of a foundation model where  the model does not require additional training. 
However, existing visual in-context learning approaches involve  training and/or use of additional task-related data. 
Further,  when we investigated V-ICL approaches  based on uncurated datasets\footnote{We use approaches based on uncurated datasets as the basis for our analysis/formulations since they do not explicitly use images from out-of-domain tasks for training and are much closer to in-context learning in the NLP community as compared to the approaches that use curated datasets.} (\eg \citet{bar2022visualprompting}) through multiple experiments, we observe the following, 
(1) the model fails to reconstruct the input prompt, despite it being unmasked in the input, as shown in the top-left image of the foundation model's output in  \cref{fig:vp_drawbacks}, 
(2) the model struggles in the presence of multiple objects in the image as shown in \cref{fig:vp_drawbacks} where it incorrectly predicts the \textit{bird feeder} as the foreground, and 
(3) when provided with the query and its groundtruth as the example prompt, the model struggles to make correct predictions despite having access to the groundtruth for the query (see \cref{fig:vp_drawbacks_toy}).
These observations indicate that the model is unable to fully leverage the prompts. We attribute this inability of the model to lack of appropriate context interpretation (1) between the source and target of the example prompt which is essential for task inference, and (2) between the query and the source of the example prompt which is essential for accurate predictions. 

To alleviate these limitations, we propose a novel inference-only Stable Diffusion based visual in-context learning pipeline (SD-VICL), that unlike existing approaches, requires no additional training. 
Stable Diffusion \citep{rombach2022stablediff} is a latent diffusion model that generates high-quality images by iteratively refining random noise through a denoising process. 
The denoising process at a given time step $t$ employs a denoising U-Net which comprises of multiple self-attention layers operating at resolutions $16\times16$, $32\times32$, and $64\times64$.  
At each self-attention layer ($l$), the input features of the intermediate noise latent, $\phi(z_t)$, are transformed to Query ($Q$), Key ($K$), and Value ($V$) vectors using linear layers.
The $Q$ and $K$ vectors are used to compute the self-attention map using:
\begin{equation} \label{eq:attn}
    \small
    \alpha^{(t)} = \text{\normalsize{softmax}} \left ( \frac{Q^{(t)} \cdot K^{{(t)}^T}}{\sqrt{d}} \right ),
\end{equation}
which captures the correspondences within the image. 
The parameter $d$ denotes the feature dimension of the $Q$ vector. 
The self-attention map ($\alpha$) is used to update the intermediate feature map using the feature update, $\Delta\phi$, which is computed as, 
\begin{equation} \label{eq:delta_phi}
    \small 
    \Delta\phi^{(t)} = \alpha^{(t)} \cdot V^{(t)}.
\end{equation}
This updated set of features, along with the updates from the cross-attention layers are used to compute the intermediate noise predictions. 
In this paper, we focus only on the self-attention computations. Hence, for simplicity, we write the iterative denoising process as
\begin{equation}
    \small
    z^{(t-1)} = z^{(t)} - \epsilon(Q^{(t)}, K^{(t)}, V^{(t)}),
\end{equation}
where the predicted noise, $\epsilon$, is a function of the $Q,K,$ and $V$ vectors of the self-attention layers.

\subsection{Repurposing Stable Diffusion for Visual In-Context Learning (SD-VICL)}\label{sec:proposed_method}

As discussed previously, for successful in-context learning, a foundation model needs to appropriately infer (1) \textit{the task}: the relationship between the prompt image ($A$) and prompt groundtruth ($B$), and (2) \textit{the context}: the relationship between the query image ($C$) and the prompt image ($A$). 
In an attempt to ensure that these relationships are effectively inferred, we formulate a novel attention computation in place of the traditional self-attention computation in the upsample layers of the denoising U-Net. 
This formulation is inspired by a few recent research \citep{cao2023masactrl,alaluf2024cross,tumanyan2023plug,patashnik2023localizing} that repurpose the self-attention layers of Stable Diffusion for tasks like image editing and style transfer. 

First, each of the input images: prompt image ($A$), prompt groundtruth ($B$), and query image ($C$) are inverted to the Stable Diffusion's noise space to obtain, $z_A^{(T)}$, $z_B^{(T)}$, and $z_C^{(T)}$ respectively, using an off-the-shelf inversion model \citep{huberman2024ddpminv}.
Then, each of these images are iteratively denoised in parallel without any changes to the default denoising pipeline as follows:
\begin{equation} \label{eq:abc_denoise}
    \small
    z_p^{(t-1)} = z_p^{(t)} - \epsilon(Q_p^{(t)}, K_p^{(t)}, V_p^{(t)}),
\end{equation}
where $p \in \{A, B, C\}, t \in [T,1], t\in \mathbb{Z}^+$.

\begin{figure*}[t]
    \centering
    \includegraphics[width=0.85\linewidth]{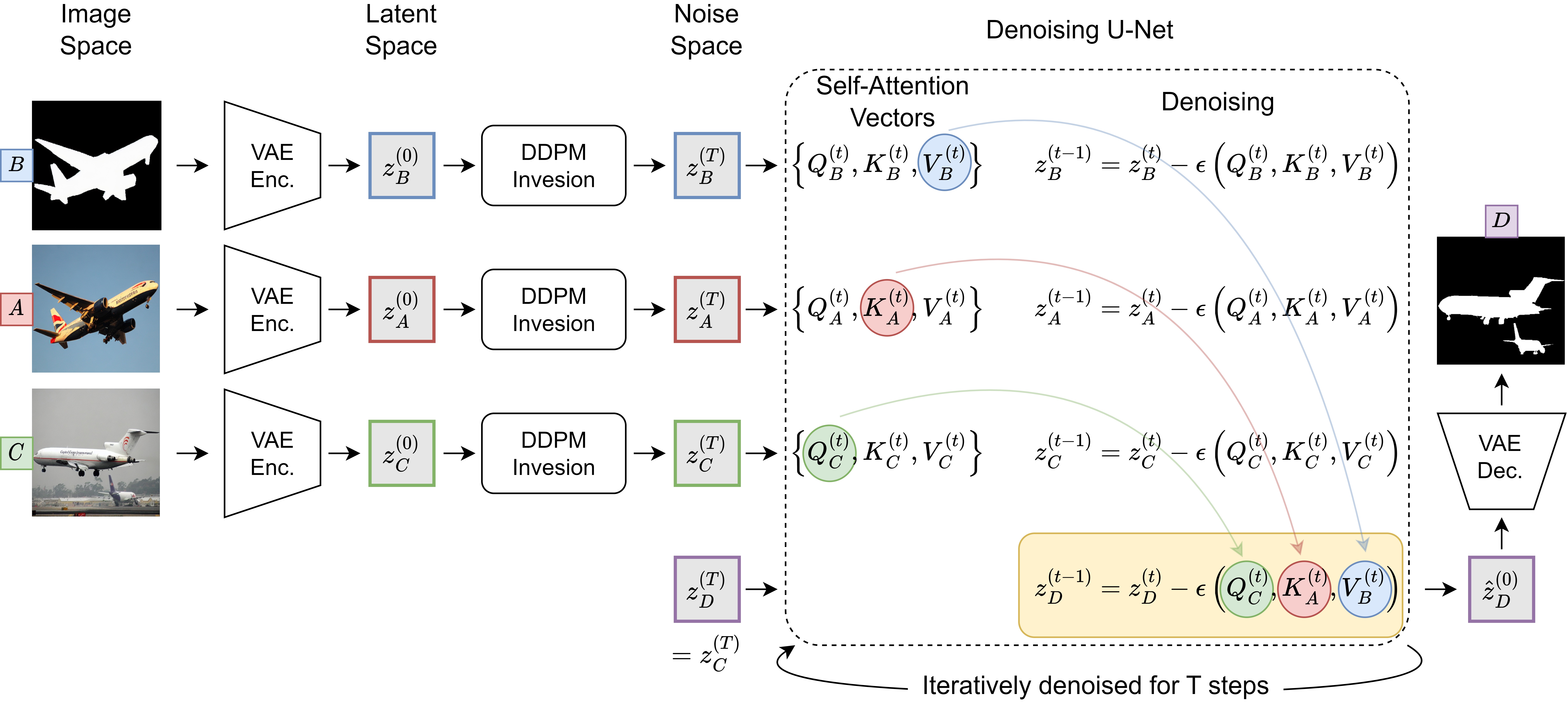}
    \caption{Latent representations of the query and prompt images from the Stable Diffusion model's VAE encoder are transformed to the noise space using \citep{huberman2024ddpminv}.
    Each noisy latent is iteratively denoised using the denoising U-Net. Prediction ($D$) path is initialized with the noise space of $C$, and at each denoising step, the computations within the self-attention layers are modified to enhance visual ICL capabilities. With the Query (from prompt image-$A$), the Key (from prompt groundtruth-$B$), and the Value (from query image-$C$) vectors, we  reformulate the intermediate feature update to infuse context between the example prompts and the query image into the foundation model.
    This process is iteratively performed for $T$ denoising steps to obtain the denoised prediction latent, which is then fed to the VAE decoder to obtain the final prediction $D$. }\vspace{-12pt}
    \label{fig:pipeline}
\end{figure*}

Since the output image is expected to be structurally similar to the query image for most vision tasks, we initialize the noise space in the prediction pipeline $D$ using the noise space of the query image (\ie, $z_D^{(T)} \leftarrow z_C^{(T)}$). 

Different from \cref{eq:abc_denoise}, the query prediction is denoised using,
\begin{equation} \label{eq:d_denoise}
    \small
    z_D^{(t-1)} = z_D^{(t)} - \epsilon(Q_C^{(t)}, K_A^{(t)}, V_B^{(t)}),
\end{equation}
where $t \in [T,1], t\in \mathbb{Z}^+$. 

Expanding the feature update formulation for $D$ we get,
\begingroup
\small
\begin{align}
    \Delta\phi_D^{(t)} &= \alpha_D^{(t)} \cdot V_B^{(t)}, \\    
    \Delta\phi_D^{(t)} &= \text{\normalsize softmax} \left ( \frac{Q_C^{(t)} \cdot K_A^{{(t)}^T}}{\tau \cdot \sqrt{d}} \right ) \cdot V_B^{(t)}. \label{eq:delta_phi_d}
\end{align}
\endgroup
In contrast to \cref{eq:attn,eq:delta_phi}, the attention map is formulated using the Query vector of the query image ($C$) and the Key vector of the prompt image ($A$). 
The $Q_C$ vector comprises of the semantics of each spatial location in $C$, and the $K_A$ vector offers the context within $A$ that each query can attend to. 
Hence, each element in the attention map, \ie, $\alpha_{D(i,j)}^t$, captures how the $i^{th}$ patch of the query image correlates with the $j^{th}$ patch of the prompt image. 
This formulation explicitly enforces the context between the query image and the prompt image. 
The additional temperature hyperparameter ($\tau$) (in \cref{eq:delta_phi_d}) for the softmax computation controls the sharpness of the correlation. 
     By adjusting $\tau$, we can modulate the model's focus on attention/correlation between $C$ and $A$. A lower $\tau$ sharpens the attention, emphasizing stronger correlation, while a higher $\tau$ smoothens the attention, allowing for a broader distribution of focus across multiple correlations.
Furthermore, we use the Value vector from the prompt groundtruth ($B$) to compute $\Delta\phi_D^{(t)}$. 
This use of value vector from a different latent,  in contrast to the standard implementation of cross-attention \citep{vaswani2017attention}, where both Key and Value vectors typically come from the same set of intermediate latents, is to facilitate the prediction to be in the same domain as that of the prompt groundtruth $B$.
This formulation, which is in-place of the computation within the standard self-attention layer of Stable Diffusion, facilitates the explicit infusion of the context between the query image and the prompt in addition to improved task inference.
However, the modification of the default self-attention computation and the use of $Q$, $K$, $V$ vectors that are extracted from three distinct images create a domain gap that needs to be addressed to enhance the quality of the prediction. 
Hence, we adapt the \textit{attention map contrasting, swap-guidance}, and \textit{AdaIN} mechanisms employed in \citep{alaluf2024cross}. 

\textit{Attention map contrasting} supplements the function of the temperature hyperparameter ($\tau$), where it enhances the focus on the relevant regions (\ie semantically similar regions) between $C$ and $A$, while attenuating the irrelevant (\ie color similarity between unrelated objects). 
Additionally, this operation adjusts the scale of the values to suit the pre-trained Stable Diffusion pipeline. 
This contrasting operation can be expressed as,
\begin{equation}
    \small
    \alpha_D^{(t)} = \mu(\alpha_D^{(t)}) + \beta \cdot \left (\alpha_D^{(t)} - \mu(\alpha_D^{(t)}) \right ),\label{eq:contrast}%
\end{equation}
where $\mu$ is the mean operation and $\beta$ is a hyperparameter that controls the scale. 

\textit{Swap-guidance} is derived from classifier-free diffusion guidance implementation proposed by \citet{ho2022classifier}. 
Given the noise prediction from the default self-attention formulation ($\eta_{default}^{(t)}$) and the noise prediction from our modified formulation ($\eta_{modified}^{(t)}$), the final noise prediction ($\eta^{(t)}$) can be expressed as,
\begin{equation}
    \small
    \eta^{(t)} = \eta_{default}^{(t)} + \frac{\gamma \cdot (T-t)}{T} \cdot \left (\eta_{modified}^{(t)} - \eta_{default}^{(t)} \right),\label{eq:swap_guidance}%
\end{equation}
where $\gamma$ is a hyperparameter controlling the scale. 
This mechanism gradually incorporates the modified noise prediction as denoising progresses, which directs the denoising process through denser regions of the Stable  Diffusion's generative pipeline, thus alleviating unwarranted artifacts. 

\textit{Adaptive instance normalization (AdaIN)}, which was proposed by \citet{huang2017adain} is used to align the color distribution between the prediction ($D$), which is initialized using the noise space of the query image ($C$), and the final groundtruth/task color space (\ie color space of $B$). AdaIN is implemented as follows,
\begingroup
\small
\begin{align}
    z_D^{(t)} &\leftarrow \text{AdaIN}(z_D^{(t)}, z_B^{(t)}), \\
    z_D^{(t)} &\leftarrow \frac{z_D^{(t)} - \mu(z_D^{(t)})}{\sigma(z_D^{(t)})} \cdot \sigma(z_B^{(t)}) + \mu(z_B^{(t)}),
\end{align}
\endgroup
where $\mu$ and $\sigma$ are mean and standard deviation operations. 

\subsection{Implicitly-Weighted Prompt Ensembling (IWPE)} 

\citet{bar2022visualprompting} and \citet{xu2023improv} demonstrate that providing the model with multiple source-target example prompts improves the prediction performance of visual in-context learning. 
These methods ensemble multiple prompts at the input level by stitching together prompt images and corresponding groundtruths to form a grid-like structure in the form of a composite image. 
This composite image is then input to the model, which is expected to leverage information from multiple prompts to make a better prediction compared to the single prompt scenario. 
However, for a fixed-resolution foundation model, as the number of source-target examples increases, the effective size of each image in the grid is reduced. 
This leads to performance deterioration, mainly because of the loss of details in the input \citep{xu2023improv}.  

To address this issue, instead of ensembling prompts in the image space through a composite input, SegGPT \citep{wang2023seggpt} proposed a feature space ensembling method. 
In this method, multiple prompts are processed in parallel and aggregated by averaging the features at the end of each attention layer.   
This approach assumes that all the example prompts are equally informative and they are uniformly weighted during the averaging process. However, such uniform weighting of prompts is sub-optimal since it prevents the model from benefiting from more relevant prompts. To this end, we propose a simple but effective ensembling method by integrating the ensembling into the attention computation, allowing the prompt patches to be implicitly weighted based on their correspondences with each query patch. Specifically, we concatenate each of the $K$ and $V$ vectors from each of the prompts prior to computing $\Delta \phi_D^{(t)}$. 
Specifically, for $n$ number of prompts, we re-write \cref{eq:delta_phi_d}:
\begin{equation}
    \footnotesize
    \Delta\phi_D^{(t)} = \text{softmax} \left ( \frac{Q_C^{(t)} \cdot 
    \left (\bigoplus_{i=1}^n K_{A_i}^{(t)} \right )^T} {\tau \cdot \sqrt{d}} \right ) \cdot \left (\bigoplus_{i=1}^n V_{B_i}^{(t)} \right ),
\end{equation}
where $\oplus$ is the concatenation operation and $i$ is the index for each prompt.

\section{Experiments and Evaluations}

To demonstrate the ability of the proposed approach to generalize across different tasks, we evaluate it on six downstream tasks. 
For all these tasks, we use the unsupervised prompt retrieval \citep{zhang2023vpr} to select the candidates for the prompt images. 
Specifically, this method chooses the nearest neighbors of the query image  as prompt candidates. The nearest neighbor retrieval is based on the  cosine similarity of CLIP's vision encoder \citep{radford2021clip} embeddings between the query image and the example prompt images. 

For a fair comparison, we use Visual Prompting \citep{bar2022visualprompting} and IMProv \citep{xu2023improv} as baselines. 
Although these approaches involve a training step, unlike \cite{wang2023painter, wang2023promptdiff}, they do not rely on curated or annotated data that come from related out-of-domain tasks. 
Additionally, since IMProv supports supplementary text guidance, we include results with and without text guidance in our comparisons. 
Further, we also evaluate and compare the effectiveness of multiple prompts. 
Please refer to the supplementary for details on datasets, evaluation metrics, and sensitivity analysis for different hyperparameters. For the results in the following subsections, we choose one set of hyperparameters that provided optimal performance for all the tasks.

\subsection{Results and Analysis}\label{ssec:results}

Quantitative metrics for all the six tasks are reported in \cref{tab:quant_1}. The proposed approach consistently outperforms existing uncurated based approaches in all the tasks. 

\noindent\textbf{Foreground segmentation}: With a single prompt, the proposed method achieves absolute improvements of {8.9\% and 3.2\%} in \textit{mean intersection over union} (mIoU) on an average across all splits\footnotemark{} in the Pascal-5i dataset \citep{shaban2017pascal5i}, compared to Visual Prompting and IMProv, respectively. 
The proposed ensembling method with five prompts using the repurposed Stable Diffusion model further improves the mIoU by {11.6\%} (in absolute terms) as compared to the single prompt case. 

\noindent\textbf{Single object detection}: 
Similar to foreground segmentation, with a single prompt, the proposed method improves the mIoU by {5.3\% and 7.1\%} (in absolute terms) on average across all splits\protect\footnotemark[\value{footnote}] in the Pascal-5i dataset \citep{shaban2017pascal5i} as compared to Visual Prompting and IMProv, respectively. 
The use of multiple prompts through implicitly-weighted prompt ensembling results in an additional {5.6\%} of improvement as compared to the single prompt case. 
\footnotetext{See supplementary for the performance on each split.}

\begin{figure*}[t!]
    \centering
    \includegraphics[width=0.82\linewidth]{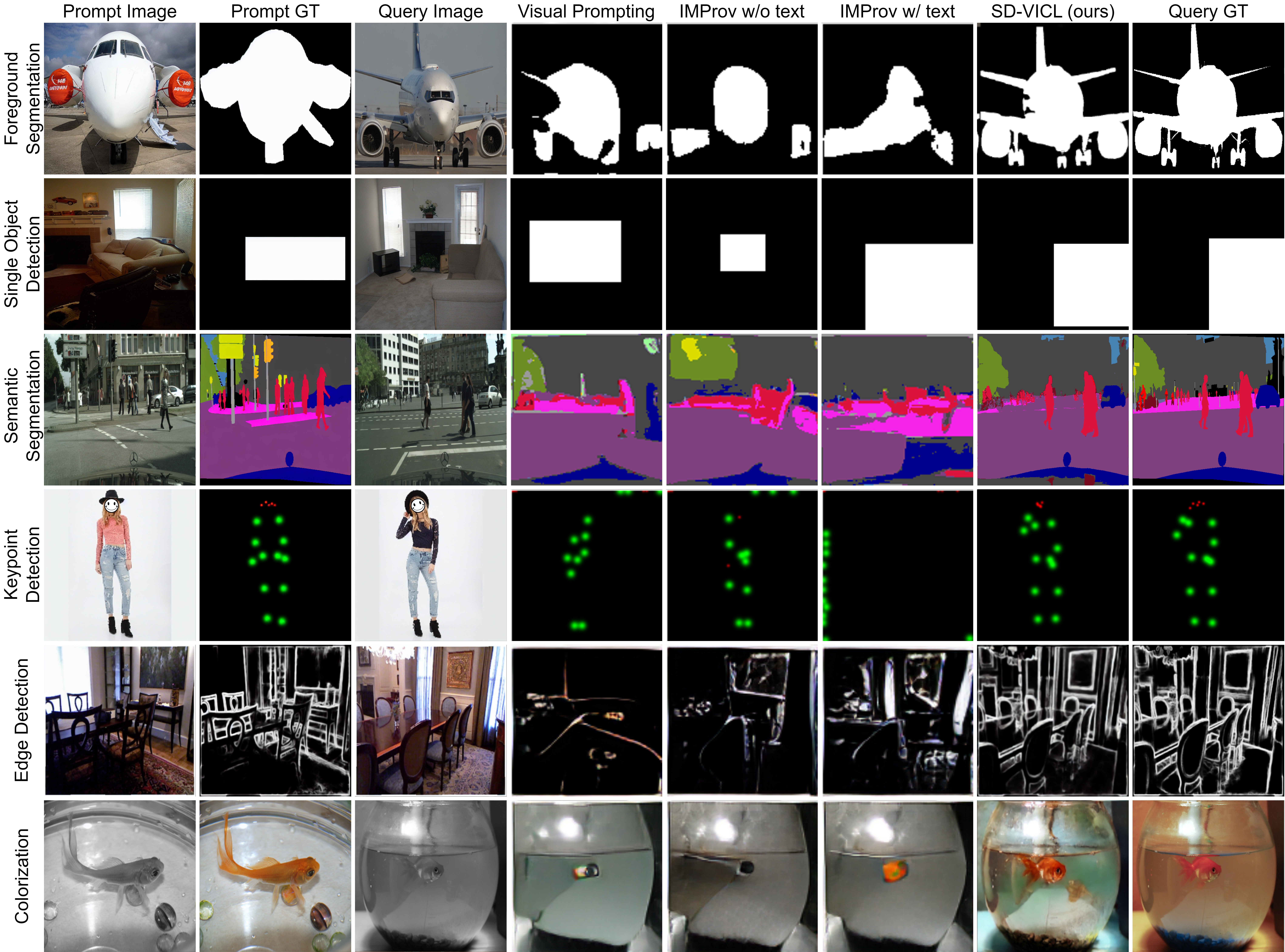}
    \caption{Qualitative evaluation, where we compare the performance of Visual Prompting \citep{bar2022visualprompting} and IMProv \citep{xu2023improv} with our proposed method on six different tasks. It can be seen that our method produces visually superior results as compared to  the  baselines.}
    \vspace{-11pt}
    \label{fig:qualitative_1}
\end{figure*}

\noindent\textbf{Semantic segmentation}: On the Cityscapes dataset \citep{cordts2016cityscapes}, our method with a single prompt results in an absolute improvement of 0.8\% in mIoU and 3.4\% increase in accuracy as compared to Visual Prompting. 
Similarly, we observe an absolute gain of 6.1\%  in mIoU and a 6.5\% in accuracy as compared to IMProv. 
With IWPE, the mIoU and accuracy further improve by 4.9\% and 6.8\% (in absolute terms) as compared to the single prompt based approach. 
The proposed approach with a single prompt in the case of semantic segmentation achieves lower performance improvements over existing methods as compared to other tasks like foreground segmentation and keypoint detection. 
However, with additional prompts, the proposed method is able to achieve much higher gains similar to what is observed in other tasks. 
This performance boost can be attributed to two key factors: (1) multiple prompts help resolve conflicts between similar or overlapping classes by mitigating  inter-class attention ambiguities, and (2) a single prompt is unlikely to contain all classes that are potentially present in the query image of a multi-class task and in such cases, access to multiple prompts increases the probability that all relevant classes are available to the model for making accurate predictions. 

\noindent\textbf{Keypoint detection}: When evaluated on the DeepFashion dataset \citep{liu2016deepfashion}, our method demonstrates a substantial improvement, reducing the MSE by ~$6\times$  while achieving a  $7\times$ gain in \textit{percentage of correct keypoints} (PCK), compared to the best-performing baseline. 
Furthermore, the use of five prompts enhances these metrics, yielding additional relative gains of 18.5\% for MSE and 6.5\% for PCK.  

\noindent\textbf{Edge detection}: On the NYUDv2 dataset \citep{silberman2012nyudv2}, our approach reduces the mean squared error (MSE) by 72.0\% and
LPIPS \citep{zhang2018lpips} by 60.6\% as compared to the best performing existing approach. Note that these metrics improve an additional 24.5\% and 21.4\%, respectively, with our prompt ensembling method.

\noindent\textbf{Colorization}: Evaluating on the ImageNet dataset \citep{russakovsky2015imagenet}, our model reduces LPIPS by 27.9\% and FID \citep{heusel2017fid} by 49.0\%, relative to IMProv, which yields the next best performance, and shows further gains of 19.1\% and 16.0\% respectively, with five prompts.

\begin{table*}[tbp]
  \centering
\resizebox{0.95\linewidth}{!}{

  \begin{tabular}{l|c|c|cc|cc|cc|cr}
    \toprule    
    Model 
    & Foreground Seg.
    & Object Det.
    & \multicolumn{2}{c}{Semantic Segmentation}
    & \multicolumn{2}{c}{Keypoint Detection}
    & \multicolumn{2}{c}{Edge Detection}
    & \multicolumn{2}{c}{Colorization} \\

    & (Pascal-5i)
    & (Pascal-5i)
    & \multicolumn{2}{c}{(Cityscapes)}
    & \multicolumn{2}{c}{(DeepFashion)}
    & \multicolumn{2}{c}{(NYUDv2)}
    & \multicolumn{2}{c}{(ImageNet)} \\

    & mIoU~$\uparrow$
    & mIoU~$\uparrow$
    & mIoU~$\uparrow$ & Acc.~$\uparrow$ 
    & MSE~$\downarrow$ & PCK~$\uparrow$
    & MSE~$\downarrow$ & LPIPS~$\downarrow$
    & LPIPS~$\downarrow$ & FID~$\downarrow$ \\
    
    \midrule
    
    \multicolumn{4}{l}{Number of Example Prompts: 1}  \vspace{2pt} \\
    
    \quad Visual Prompting \citep{bar2022visualprompting}
    & 35.04
    & \underline{46.29}
    & \underline{21.30} & \underline{71.52}
    & 35.83 & 10.87
    & \underline{0.1006} & \underline{0.3925}
    & 0.4166 & 111.06 \\
                         
    \quad IMProv (w/o text) \citep{xu2023improv}
    & 39.50
    & 43.82
    & 15.97 & 68.44
    & \underline{31.30} & \underline{18.38}
    & 0.1059 & 0.4278
    & \underline{0.3895} & \underline{104.74} \\                            
    
    \quad IMProv (w/ text) \citep{xu2023improv}
    & \underline{40.69}
    & 44.42
    & 15.80 & 67.88
    & 44.67 & 2.81 
    & 0.1125 & 0.5146
    & 0.3899 & 105.90 \\

    \quad SD-VICL (ours) 
    & \textbf{43.92}
    & \textbf{51.55}
    & \textbf{22.08} & \textbf{74.93}
    & \textbf{5.36} & \textbf{77.19}  
    & \textbf{0.0282} & \textbf{0.1548}
    & \textbf{0.2810} & \textbf{53.40} \\

    \midrule

    \multicolumn{6}{l}{Number of Example Prompts: 5} \vspace{2pt} \\
    
    \quad Visual Prompting 
    & 36.37
    & 48.09
    & 21.70 & 70.95
    & 36.57 & 10.32 
    & 0.1000 & 0.3900
    & 0.4156 & 112.91 \\
                               
    \quad SD-VICL (ours)
    & \textbf{55.49}
    & \textbf{57.10}
    & \textbf{27.01} & \textbf{81.75}
    & \textbf{4.37} & \textbf{82.24}
    &  \textbf{0.0213} & \textbf{0.1216}
    & \textbf{0.2272} & \textbf{44.84} \\ 

    \bottomrule
  \end{tabular}
  }
  \vspace{-5pt}
  \caption{Quantitative performance comparison of the proposed approach with recent approaches on foreground segmentation, single object detection, semantic segmentation, keypoint detection, edge detection, and colorization.
  }
  \label{tab:quant_1}
\end{table*}

\begin{table*}[tbp]
    \centering
    \begin{minipage}{0.54\textwidth}
        \centering
        \resizebox{\linewidth}{!}{
            \begin{tabular}{l|c|c|cc|cc}
                \toprule    
                Model 
                & \multicolumn{1}{c|}{FG Seg.}
                & \multicolumn{1}{c|}{Obj. Det.}
                & \multicolumn{2}{c|}{Edge Det.}
                & \multicolumn{2}{c}{Colorization} \\
            
                & mIoU~$\uparrow$ 
                & mIoU~$\uparrow$ 
                & MSE~$\downarrow$ & LPIPS~$\downarrow$
                & LPIPS~$\downarrow$ & FID~$\downarrow$ \\
                
                \midrule
                
                Painter \citep{wang2023painter}
                & 55.09 
                & 54.28  
                & 0.0926 & 0.7294
                & 0.3474 & 64.16 \\
            
                LVM \citep{bai2024lvm}
                & 50.98 
                & 52.67  
                & 0.0499 & 0.4259
                & 0.3142 & 56.40 \\

                Prompt Diffusion \citep{wang2023promptdiff}
                & 17.53 
                & 26.86  
                & 0.0430 & 0.3311
                & 0.6997 & 98.91 \\
            
                SD-VICL (ours)
                & \textbf{55.49} 
                & \textbf{57.10} 
                &  \textbf{0.0213} & \textbf{0.1216}
                & \textbf{0.2272} & \textbf{44.84} \\   
            
                \bottomrule
            \end{tabular}
        }
        \caption{Extended quantitative evaluations against V-ICL models that train on task-related data.}\label{tab:quant_painter_lvm}
    \end{minipage}
    \hfill
    \begin{minipage}{0.4\textwidth}
    \centering
        \resizebox{\linewidth}{!}{
            \begin{tabular}{l|cc|cc}
                \toprule    
                Model 
                & \multicolumn{2}{c|}{Sem. Seg.}
                & \multicolumn{2}{c}{Keypoint Det.} \\
            
                & mIoU~$\uparrow$ & Acc.~$\uparrow$
                & MSE~$\downarrow$ & PCK~$\uparrow$ \\
                
                \midrule
                
                DAS \citep{tian2024diffuse}
                & 21.2 & 76.0
                & \_ & \_ \\
            
                StableKeypoints \citep{hedlin2024stablekeypoints}
                & \_ & \_
                & 6.46 & 70.0 \\
            
                SD-VICL (ours)
                &  \textbf{31.52} & \textbf{82.72}
                & \textbf{4.37} & \textbf{82.24} \\   
            
                \bottomrule
            \end{tabular}
        }
        \caption{Comparison with task-specific models based on Stable Diffusion.} \label{tab:quant_taskspecific}
    \end{minipage}
    \vspace{-12pt}
\end{table*}
 
Overall, our method outperforms both Visual Prompting and IMProv across all tasks by considerable margins.
Additionally, the integration of multiple prompts using IWPE consistently yields performance improvements over the single-prompt scenario. 

Supplementing the quantitative results, we also demonstrate visual comparisons of the outputs for each task in \cref{fig:qualitative_1}\footnote{See supplementary for additional qualitative results for each task.}, further highlighting the superior performance of our method over baselines. We observe that while  both Visual Prompting and IMProv perform reasonably in single object detection and foreground segmentation, they show a weaker performance in tasks such as colorization, edge detection, and semantic segmentation, where capturing fine details and preserving structural information of the query is crucial.  
Moreover, both baselines notably struggle with keypoint detection, whereas the proposed approach demonstrates far superior results, indicating its robustness to different out-of-domain tasks.

\subsection{Additional Evaluations} \label{sec:add_eval}

\noindent\textbf{Comparison with V-ICL models trained on task-related data:}
To provide a comprehensive evaluation, we compare our training-free approach against prior methods trained explicitly on task-related data, including Painter \citep{wang2023painter}, LVM \citep{bai2024lvm}, and Prompt Diffusion \citep{wang2023promptdiff}. 
Quantitative results are presented in \cref{tab:quant_painter_lvm}, and qualitative comparisons are shown in Fig. 7 (supplementary).  
Despite the lack of any explicit training, our method consistently outperforms all three baselines across multiple vision tasks.
Analyzing the results, we observe that all three models tend to overfit to their training tasks, leading to poor generalization when presented with novel tasks.
For instance, Painter incorrectly outputs a depth map instead of edges, despite the prompt pair defining the task has an edge map as the target (Fig. 7 in supplementary).
Further, while Painter and LVM perform well on tasks, foreground segmentation and object detection, when the query image consists of only a single class, they struggle in multi-class scenarios, often segmenting all classes instead of the intended one (Fig. 8a in supplementary).
Additionally, LVM exhibits inconsistencies in its outputs, where for a given task, despite the format/domain of the inputs remaining unchanged, we observe that the generated output belongs to diverse domains (Fig. 8b in supplementary). 
Moreover, Prompt Diffusion performs notably poorly on most tasks, often generating incorrect outputs, with the exception of edge detection --- a task it was exposed to during training. 
Ideally, a V-ICL model should infer context and task from the input, but all three models fall short in this aspect. 
In contrast, our training-free method demonstrates superior generalization and effective context and task inference as intended by V-ICL, underscoring the advantages of uncovering V-ICL properties without additional training.
For a detailed discussion and visual examples, we refer the reader to Sec. E in the supplementary.

\noindent\textbf{Comparison with task-specific models based on Stable Diffusion:}
Here, we compare our method against approaches that adapt Stable Diffusion to facilitate specific tasks, such as semantic segmentation and keypoint detection.
DAS \citep{tian2024diffuse} is one such work that proposes a pipeline that aggregates the self-attention layers of Stable Diffusion iteratively to predict segmentation masks for each class facilitating semantic segmentation.
Similarly, StableKeypoints \citep{hedlin2024stablekeypoints} optimizes on the text embeddings such that cross-attention maps condense to landmarks in a given image facilitating keypoint detection. 
We present the performance comparisons against these task-specific models in \cref{tab:quant_taskspecific}, where our approach outperforms both \citep{tian2024diffuse} and \citep{hedlin2024stablekeypoints}.
For semantic segmentation, we follow the evaluation protocol of DAS  (\ie Hungarian matching) to ensure a fair comparison and achieve absolute improvements of 10.3\% in mIoU and 6.7\% in accuracy. 
In keypoint detection, we surpass StableKeypoints with a 32.4\% reduction in MSE and 12.2\% improvement in PCK.
While DAS and StableKeypoints achieve strong results in their respective tasks, they remain constrained by their task-specific nature. 
In contrast, our method not only delivers superior performance but also generalizes across diverse tasks without requiring any modification in the inference.
Furthermore, unlike these methods, which derive predictions from intermediate feature or attention maps, our model directly utilizes the generative process to produce outputs, making it inherently more flexible and effective. 

\subsection{Ablation Study} \label{sec:abl}

In this section, we present the details of various ablations conducted to demonstrate the significance of key design choices. 
Specifically, we evaluate the effectiveness of implicitly-weighted prompt ensembling and examine the impact of number of prompts on performance. 
For additional ablations, including alternative attention formulations and sensitivity analyses on hyperparameters such as attention temperature, self-attention resolution, contrastive strength, swap-guidance scale, and AdaIN, we refer the reader to Sec. F in the supplementary.

\noindent\textbf{Effectiveness of implicitly-weighted prompt ensembling (IWPE):}
Here, we compare our proposed ensembling method, 
IWPE, to SegGPT's feature ensembling (FE) approach. For all tasks except semantic segmentation, we observe that both methods yield similar gains over the single prompt scenario. In the case of semantic segmentation, IWPE demonstrates significantly better performance.
As tabulated in \cref{tab:prompt_ens}, while FE yields an absolute improvement of merely 0.9\% in mIoU and 3.9\% in accuracy, IWPE achieves substantial gains, improving mIoU by 4.9\% and accuracy by 6.8\% over the single prompt case. This notable improvement is primarily due to the capability of our ensembling method to dynamically weigh each prompt based on its correspondence with the query image.
In contrast, as discussed previously, the feature ensembling approach assumes that all prompts are equally informative, which is sub-optimal, especially in the case of tasks involving multiple classes like semantic segmentation. 
\begin{table}[tbp!]
\centering
\resizebox{0.68\linewidth}{!}{
\begin{tabular}{c|c|c c}
\toprule
\# Prompts & Ensembling   & mIoU~$\uparrow$  & Acc.~$\uparrow$\\ 
\midrule
1  & N/A  & 22.08 & 74.93 \\ 
5  & FE   & 22.97 & 78.84 \\ 
5  & IWPE & \textbf{27.01} & \textbf{81.75} \\ 
\bottomrule
\end{tabular}
}
\vspace{-5pt}
\caption{Ablation of prompt-ensembling evaluated on semantic segmentation.}
\label{tab:prompt_ens}
\vspace{-12pt}
\end{table}
For instance, one prompt may provide a stronger correlation with a specific class, while another prompt, would have a weaker association with that class. 
Our method by implicitly weighing the influence of different prompts based on their relative correspondence to the query, better captures these nuances, resulting in a significant boost in performance compared to SegGPT's uniformly weighted feature ensembling. 
Therefore, while both FE and IWPE perform at par in simpler tasks where the prompts are equally informative, our method is more beneficial for tasks that involve complex multi-class structures.

\begin{figure}[t]
    \centering
    \scalebox{1.0}{
    \begin{minipage}{\linewidth}
    \begin{subfigure}{0.49\linewidth}
        \centering
        \includegraphics[width=\linewidth]{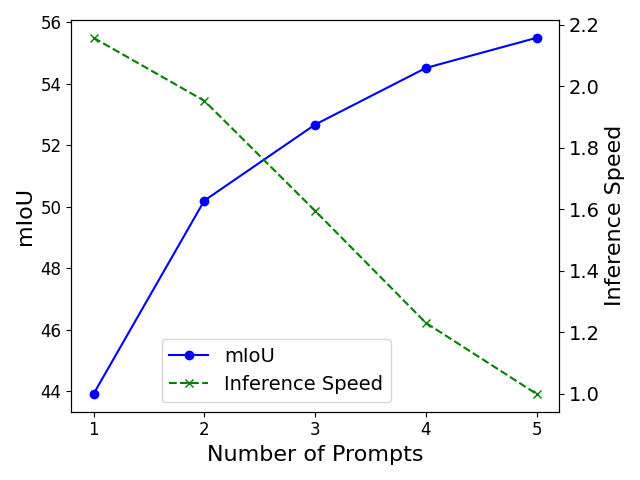}
        \caption{}
        \label{fig:abl_prompts_perf}
    \end{subfigure}
    \hfill
    \begin{subfigure}{0.5\linewidth}
        \centering
        \includegraphics[width=\linewidth]{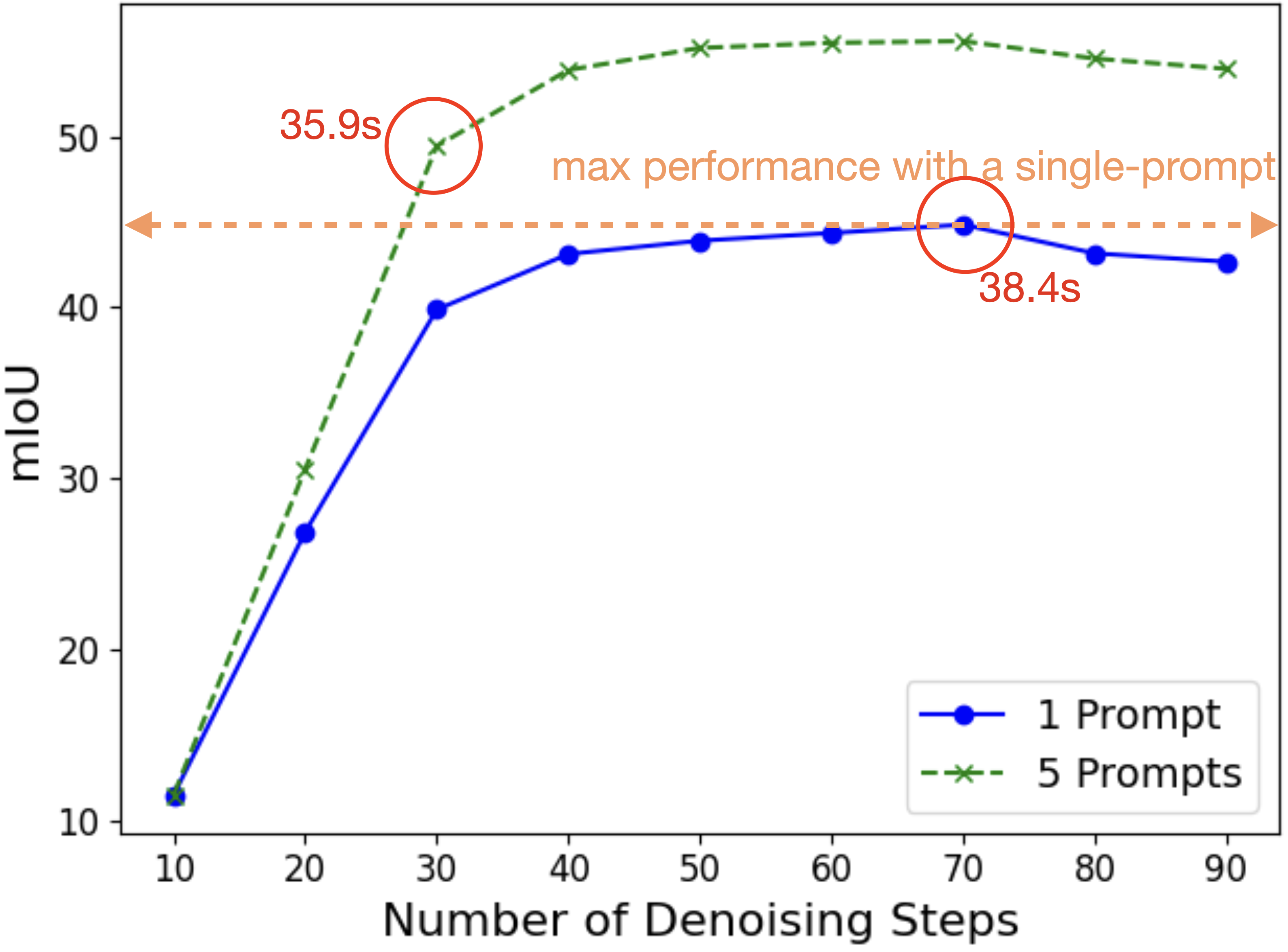}
        \caption{}
        \label{fig:abl_prompts_steps}
    \end{subfigure}
    \end{minipage}
    }
    \captionsetup{singlelinecheck=off}
    \vspace{-8pt}
    \caption{Effect of the number of prompts on the performance and inference speed\protect\footnotemark[\value{footnote}]. 
    While (a) depicts the variation in mIoU and the inference speed, with the number of prompts,  (b) illustrates the variation in mIoU with the number of denoising steps for both single and five prompt cases. 
    }\vspace{-12pt}
\end{figure}

\noindent\textbf{Impact of number of prompts:}
Here, we investigate the effect of the number of prompts on the performance and the inference speed. 
As illustrated in \cref{fig:abl_prompts_perf}, the mIoU scores improve with an increasing number of prompts.
This result can be attributed to the model's ability to effectively leverage multiple prompts to better infer the task and establish the correspondence between the query and the prompts. 
The additional prompts help  reduce ambiguities that may arise from using a single prompt. 
However, these gains come at the cost of reduced inference speed; specifically using five prompts cuts the inference speed to approximately half that of a single prompt. 
Besides the prompts, the number of denoising steps in the Stable Diffusion model is a key factor impacting the inference speed. 
This led us to the question --- \textit{With the availability of multiple prompts, does the model require as many denoising steps as in the single prompt case?} 
In other words, we explored if we can trade-off a few denoising steps without impacting the performance when additional context is provided  by multiple prompts. 
To this end, we conducted experiments where we evaluated the SD-VICL approach with different  denoising steps. 
As shown in  \cref{fig:abl_prompts_steps}, the mIoU for the five-prompt scenario at 30 denoising steps surpasses the maximum performance of the single-prompt scenario. Moreover, the maximum performance for the single-prompt case, achieved with 70 denoising steps, requires 38.4 seconds of inference time. In contrast,  the five-prompt case achieves higher performance in just 35.9 seconds with 30 denoising steps\footnotemark{}.

\footnotetext{The inference times are based on evaluations on a single A100 GPU.} 

\section{Conclusion}

In this paper, we propose a novel pipeline for visual in-context learning that explicitly incorporates the context between the query image and the prompt. 
Unlike existing visual in-context learning methods which rely on training or fine-tuning foundation models on curated or uncurated data to enable in-context learning, our approach is entirely training-free. 
Specifically, we introduce an in-place attention re-computation within the self-attention layers of an off-the-shelf Stable Diffusion model. 
Additionally, the proposed implicitly-weighted prompt ensembling technique facilitates effective integration of context through multiple prompts by implicitly weighing the prompts based on the relative correspondences. 
The versatility of the proposed method is demonstrated by its successful generalization across six different tasks, where it outperforms both V-ICL models trained on uncurated and task-related data as well as task-specific models based on Stable Diffusion. 
Crucially, our method achieves these results without requiring any additional training or annotated task-specific datasets, highlighting its ability to uncover and leverage the V-ICL capabilities of Stable Diffusion.
{
    \small
    \bibliographystyle{ieeenat_fullname}
    \bibliography{main}

\begin{thebibliography}{48}
\providecommand{\natexlab}[1]{#1}
\providecommand{\url}[1]{\texttt{#1}}
\expandafter\ifx\csname urlstyle\endcsname\relax
  \providecommand{\doi}[1]{doi: #1}\else
  \providecommand{\doi}{doi: \begingroup \urlstyle{rm}\Url}\fi

\bibitem[Alaluf et~al.(2024)Alaluf, Garibi, Patashnik, Averbuch-Elor, and Cohen-Or]{alaluf2024cross}
Yuval Alaluf, Daniel Garibi, Or Patashnik, Hadar Averbuch-Elor, and Daniel Cohen-Or.
\newblock Cross-image attention for zero-shot appearance transfer.
\newblock In \emph{ACM SIGGRAPH 2024 Conference Papers}, pages 1--12, 2024.

\bibitem[Bai et~al.(2024)Bai, Geng, Mangalam, Bar, Yuille, Darrell, Malik, and Efros]{bai2024lvm}
Yutong Bai, Xinyang Geng, Karttikeya Mangalam, Amir Bar, Alan~L Yuille, Trevor Darrell, Jitendra Malik, and Alexei~A Efros.
\newblock Sequential modeling enables scalable learning for large vision models.
\newblock In \emph{Proceedings of the IEEE/CVF Conference on Computer Vision and Pattern Recognition}, pages 22861--22872, 2024.

\bibitem[Bar et~al.(2022)Bar, Gandelsman, Darrell, Globerson, and Efros]{bar2022visualprompting}
Amir Bar, Yossi Gandelsman, Trevor Darrell, Amir Globerson, and Alexei Efros.
\newblock Visual prompting via image inpainting.
\newblock \emph{Advances in Neural Information Processing Systems}, 35:\penalty0 25005--25017, 2022.

\bibitem[Brooks et~al.(2023)Brooks, Holynski, and Efros]{brooks2023instructpix2pix}
Tim Brooks, Aleksander Holynski, and Alexei~A Efros.
\newblock Instructpix2pix: Learning to follow image editing instructions.
\newblock In \emph{Proceedings of the IEEE/CVF conference on computer vision and pattern recognition}, pages 18392--18402, 2023.

\bibitem[Brown et~al.(2020)Brown, Mann, Ryder, Subbiah, Kaplan, Dhariwal, Neelakantan, Shyam, Sastry, Askell, et~al.]{brown2020gpt3}
Tom Brown, Benjamin Mann, Nick Ryder, Melanie Subbiah, Jared~D Kaplan, Prafulla Dhariwal, Arvind Neelakantan, Pranav Shyam, Girish Sastry, Amanda Askell, et~al.
\newblock Language models are few-shot learners.
\newblock \emph{Advances in neural information processing systems}, 33:\penalty0 1877--1901, 2020.

\bibitem[Cao et~al.(2023)Cao, Wang, Qi, Shan, Qie, and Zheng]{cao2023masactrl}
Mingdeng Cao, Xintao Wang, Zhongang Qi, Ying Shan, Xiaohu Qie, and Yinqiang Zheng.
\newblock Masactrl: Tuning-free mutual self-attention control for consistent image synthesis and editing.
\newblock In \emph{Proceedings of the IEEE/CVF International Conference on Computer Vision}, pages 22560--22570, 2023.

\bibitem[Chowdhery et~al.(2023)Chowdhery, Narang, Devlin, Bosma, Mishra, Roberts, Barham, Chung, Sutton, Gehrmann, et~al.]{chowdhery2023palm}
Aakanksha Chowdhery, Sharan Narang, Jacob Devlin, Maarten Bosma, Gaurav Mishra, Adam Roberts, Paul Barham, Hyung~Won Chung, Charles Sutton, Sebastian Gehrmann, et~al.
\newblock Palm: Scaling language modeling with pathways.
\newblock \emph{Journal of Machine Learning Research}, 24\penalty0 (240):\penalty0 1--113, 2023.

\bibitem[Cordts et~al.(2016)Cordts, Omran, Ramos, Rehfeld, Enzweiler, Benenson, Franke, Roth, and Schiele]{cordts2016cityscapes}
Marius Cordts, Mohamed Omran, Sebastian Ramos, Timo Rehfeld, Markus Enzweiler, Rodrigo Benenson, Uwe Franke, Stefan Roth, and Bernt Schiele.
\newblock The cityscapes dataset for semantic urban scene understanding.
\newblock In \emph{Proc. of the IEEE Conference on Computer Vision and Pattern Recognition (CVPR)}, 2016.

\bibitem[Dosovitskiy et~al.(2021)Dosovitskiy, Beyer, Kolesnikov, Weissenborn, Zhai, Unterthiner, Dehghani, Minderer, Heigold, Gelly, Uszkoreit, and Houlsby]{dosovitskiy2020vit}
Alexey Dosovitskiy, Lucas Beyer, Alexander Kolesnikov, Dirk Weissenborn, Xiaohua Zhai, Thomas Unterthiner, Mostafa Dehghani, Matthias Minderer, Georg Heigold, Sylvain Gelly, Jakob Uszkoreit, and Neil Houlsby.
\newblock An image is worth 16x16 words: Transformers for image recognition at scale.
\newblock In \emph{International Conference on Learning Representations}, 2021.

\bibitem[Esser et~al.(2021)Esser, Rombach, and Ommer]{esser2021vqgan}
Patrick Esser, Robin Rombach, and Bjorn Ommer.
\newblock Taming transformers for high-resolution image synthesis.
\newblock In \emph{Proceedings of the IEEE/CVF conference on computer vision and pattern recognition}, pages 12873--12883, 2021.

\bibitem[Fang et~al.(2024)Fang, Li, Li, Buhmann, Loy, and Liu]{fang2024point}
Zhongbin Fang, Xiangtai Li, Xia Li, Joachim~M Buhmann, Chen~Change Loy, and Mengyuan Liu.
\newblock Explore in-context learning for 3d point cloud understanding.
\newblock \emph{Advances in Neural Information Processing Systems}, 36, 2024.

\bibitem[Geng and Liu(2023)]{openlm2023openllama}
Xinyang Geng and Hao Liu.
\newblock Openllama: An open reproduction of llama, 2023.

\bibitem[Hao et~al.(2022)Hao, Song, Dong, Huang, Chi, Wang, Ma, and Wei]{hao2022language}
Yaru Hao, Haoyu Song, Li Dong, Shaohan Huang, Zewen Chi, Wenhui Wang, Shuming Ma, and Furu Wei.
\newblock Language models are general-purpose interfaces.
\newblock \emph{arXiv preprint arXiv:2206.06336}, 2022.

\bibitem[He et~al.(2022)He, Chen, Xie, Li, Doll{\'a}r, and Girshick]{he2022mae}
Kaiming He, Xinlei Chen, Saining Xie, Yanghao Li, Piotr Doll{\'a}r, and Ross Girshick.
\newblock Masked autoencoders are scalable vision learners.
\newblock In \emph{Proceedings of the IEEE/CVF conference on computer vision and pattern recognition}, pages 16000--16009, 2022.

\bibitem[Hedlin et~al.(2024)Hedlin, Sharma, Mahajan, He, Isack, Kar, Rhodin, Tagliasacchi, and Yi]{hedlin2024stablekeypoints}
Eric Hedlin, Gopal Sharma, Shweta Mahajan, Xingzhe He, Hossam Isack, Abhishek Kar, Helge Rhodin, Andrea Tagliasacchi, and Kwang~Moo Yi.
\newblock Unsupervised keypoints from pretrained diffusion models.
\newblock In \emph{Proceedings of the IEEE/CVF Conference on Computer Vision and Pattern Recognition}, pages 22820--22830, 2024.

\bibitem[Heusel et~al.(2017)Heusel, Ramsauer, Unterthiner, Nessler, and Hochreiter]{heusel2017fid}
Martin Heusel, Hubert Ramsauer, Thomas Unterthiner, Bernhard Nessler, and Sepp Hochreiter.
\newblock Gans trained by a two time-scale update rule converge to a local nash equilibrium.
\newblock \emph{Advances in neural information processing systems}, 30, 2017.

\bibitem[Ho and Salimans(2022)]{ho2022classifier}
Jonathan Ho and Tim Salimans.
\newblock Classifier-free diffusion guidance.
\newblock \emph{arXiv preprint arXiv:2207.12598}, 2022.

\bibitem[Huang and Belongie(2017)]{huang2017adain}
Xun Huang and Serge Belongie.
\newblock Arbitrary style transfer in real-time with adaptive instance normalization.
\newblock In \emph{Proceedings of the IEEE international conference on computer vision}, pages 1501--1510, 2017.

\bibitem[Huberman-Spiegelglas et~al.(2024)Huberman-Spiegelglas, Kulikov, and Michaeli]{huberman2024ddpminv}
Inbar Huberman-Spiegelglas, Vladimir Kulikov, and Tomer Michaeli.
\newblock An edit friendly ddpm noise space: Inversion and manipulations.
\newblock In \emph{Proceedings of the IEEE/CVF Conference on Computer Vision and Pattern Recognition}, pages 12469--12478, 2024.

\bibitem[Li et~al.(2020)Li, Wei, Chen, Tai, and Tang]{li2020fss}
Xiang Li, Tianhan Wei, Yau~Pun Chen, Yu-Wing Tai, and Chi-Keung Tang.
\newblock Fss-1000: A 1000-class dataset for few-shot segmentation.
\newblock In \emph{Proceedings of the IEEE/CVF conference on computer vision and pattern recognition}, pages 2869--2878, 2020.

\bibitem[Lin et~al.(2014)Lin, Maire, Belongie, Hays, Perona, Ramanan, Doll{\'a}r, and Zitnick]{lin2014coco}
Tsung-Yi Lin, Michael Maire, Serge Belongie, James Hays, Pietro Perona, Deva Ramanan, Piotr Doll{\'a}r, and C~Lawrence Zitnick.
\newblock Microsoft coco: Common objects in context.
\newblock In \emph{Computer Vision--ECCV 2014: 13th European Conference, Zurich, Switzerland, September 6-12, 2014, Proceedings, Part V 13}, pages 740--755. Springer, 2014.

\bibitem[Liu et~al.(2023{\natexlab{a}})Liu, Shen, Pun, and Cun]{liu2023explicit}
Weihuang Liu, Xi Shen, Chi-Man Pun, and Xiaodong Cun.
\newblock Explicit visual prompting for low-level structure segmentations.
\newblock In \emph{Proceedings of the IEEE/CVF Conference on Computer Vision and Pattern Recognition}, pages 19434--19445, 2023{\natexlab{a}}.

\bibitem[Liu et~al.(2023{\natexlab{b}})Liu, Zhang, Ma, Peng, et~al.]{liu2023instaflow}
Xingchao Liu, Xiwen Zhang, Jianzhu Ma, Jian Peng, et~al.
\newblock Instaflow: One step is enough for high-quality diffusion-based text-to-image generation.
\newblock In \emph{The Twelfth International Conference on Learning Representations}, 2023{\natexlab{b}}.

\bibitem[Liu et~al.(2016)Liu, Luo, Qiu, Wang, and Tang]{liu2016deepfashion}
Ziwei Liu, Ping Luo, Shi Qiu, Xiaogang Wang, and Xiaoou Tang.
\newblock Deepfashion: Powering robust clothes recognition and retrieval with rich annotations.
\newblock In \emph{Proceedings of the IEEE conference on computer vision and pattern recognition}, pages 1096--1104, 2016.

\bibitem[Patashnik et~al.(2023)Patashnik, Garibi, Azuri, Averbuch-Elor, and Cohen-Or]{patashnik2023localizing}
Or Patashnik, Daniel Garibi, Idan Azuri, Hadar Averbuch-Elor, and Daniel Cohen-Or.
\newblock Localizing object-level shape variations with text-to-image diffusion models.
\newblock In \emph{Proceedings of the IEEE/CVF International Conference on Computer Vision}, pages 23051--23061, 2023.

\bibitem[Radford et~al.(2021)Radford, Kim, Hallacy, Ramesh, Goh, Agarwal, Sastry, Askell, Mishkin, Clark, et~al.]{radford2021clip}
Alec Radford, Jong~Wook Kim, Chris Hallacy, Aditya Ramesh, Gabriel Goh, Sandhini Agarwal, Girish Sastry, Amanda Askell, Pamela Mishkin, Jack Clark, et~al.
\newblock Learning transferable visual models from natural language supervision.
\newblock In \emph{International conference on machine learning}, pages 8748--8763. PMLR, 2021.

\bibitem[Rae et~al.(2021)Rae, Borgeaud, Cai, Millican, Hoffmann, Song, Aslanides, Henderson, Ring, Young, et~al.]{rae2021gopher}
Jack~W Rae, Sebastian Borgeaud, Trevor Cai, Katie Millican, Jordan Hoffmann, Francis Song, John Aslanides, Sarah Henderson, Roman Ring, Susannah Young, et~al.
\newblock Scaling language models: Methods, analysis \& insights from training gopher.
\newblock \emph{arXiv preprint arXiv:2112.11446}, 2021.

\bibitem[Rombach et~al.(2022)Rombach, Blattmann, Lorenz, Esser, and Ommer]{rombach2022stablediff}
Robin Rombach, Andreas Blattmann, Dominik Lorenz, Patrick Esser, and Bj{\"o}rn Ommer.
\newblock High-resolution image synthesis with latent diffusion models.
\newblock In \emph{Proceedings of the IEEE/CVF conference on computer vision and pattern recognition}, pages 10684--10695, 2022.

\bibitem[Russakovsky et~al.(2015)Russakovsky, Deng, Su, Krause, Satheesh, Ma, Huang, Karpathy, Khosla, Bernstein, et~al.]{russakovsky2015imagenet}
Olga Russakovsky, Jia Deng, Hao Su, Jonathan Krause, Sanjeev Satheesh, Sean Ma, Zhiheng Huang, Andrej Karpathy, Aditya Khosla, Michael Bernstein, et~al.
\newblock Imagenet large scale visual recognition challenge.
\newblock \emph{International journal of computer vision}, 115:\penalty0 211--252, 2015.

\bibitem[Schuhmann et~al.(2022)Schuhmann, Beaumont, Vencu, Gordon, Wightman, Cherti, Coombes, Katta, Mullis, Wortsman, et~al.]{schuhmann2022laion}
Christoph Schuhmann, Romain Beaumont, Richard Vencu, Cade Gordon, Ross Wightman, Mehdi Cherti, Theo Coombes, Aarush Katta, Clayton Mullis, Mitchell Wortsman, et~al.
\newblock Laion-5b: An open large-scale dataset for training next generation image-text models.
\newblock \emph{Advances in Neural Information Processing Systems}, 35:\penalty0 25278--25294, 2022.

\bibitem[Shaban et~al.(2017)Shaban, Bansal, Liu, Essa, and Boots]{shaban2017pascal5i}
Amirreza Shaban, Shray Bansal, Zhen Liu, Irfan Essa, and Byron Boots.
\newblock One-shot learning for semantic segmentation.
\newblock \emph{arXiv preprint arXiv:1709.03410}, 2017.

\bibitem[Silberman et~al.(2012)Silberman, Hoiem, Kohli, and Fergus]{silberman2012nyudv2}
Nathan Silberman, Derek Hoiem, Pushmeet Kohli, and Rob Fergus.
\newblock Indoor segmentation and support inference from rgbd images.
\newblock In \emph{Computer Vision--ECCV 2012: 12th European Conference on Computer Vision, Florence, Italy, October 7-13, 2012, Proceedings, Part V 12}, pages 746--760. Springer, 2012.

\bibitem[Thoppilan et~al.(2022)Thoppilan, De~Freitas, Hall, Shazeer, Kulshreshtha, Cheng, Jin, Bos, Baker, Du, et~al.]{thoppilan2022lamda}
Romal Thoppilan, Daniel De~Freitas, Jamie Hall, Noam Shazeer, Apoorv Kulshreshtha, Heng-Tze Cheng, Alicia Jin, Taylor Bos, Leslie Baker, Yu Du, et~al.
\newblock Lamda: Language models for dialog applications.
\newblock \emph{arXiv preprint arXiv:2201.08239}, 2022.

\bibitem[Tian et~al.(2024)Tian, Aggarwal, Colaco, Kira, and Gonzalez-Franco]{tian2024diffuse}
Junjiao Tian, Lavisha Aggarwal, Andrea Colaco, Zsolt Kira, and Mar Gonzalez-Franco.
\newblock Diffuse attend and segment: Unsupervised zero-shot segmentation using stable diffusion.
\newblock In \emph{Proceedings of the IEEE/CVF Conference on Computer Vision and Pattern Recognition}, pages 3554--3563, 2024.

\bibitem[Touvron et~al.(2023)Touvron, Lavril, Izacard, Martinet, Lachaux, Lacroix, Rozi{\`e}re, Goyal, Hambro, Azhar, et~al.]{touvron2023llama}
Hugo Touvron, Thibaut Lavril, Gautier Izacard, Xavier Martinet, Marie-Anne Lachaux, Timoth{\'e}e Lacroix, Baptiste Rozi{\`e}re, Naman Goyal, Eric Hambro, Faisal Azhar, et~al.
\newblock Llama: Open and efficient foundation language models.
\newblock \emph{arXiv preprint arXiv:2302.13971}, 2023.

\bibitem[Tumanyan et~al.(2023)Tumanyan, Geyer, Bagon, and Dekel]{tumanyan2023plug}
Narek Tumanyan, Michal Geyer, Shai Bagon, and Tali Dekel.
\newblock Plug-and-play diffusion features for text-driven image-to-image translation.
\newblock In \emph{Proceedings of the IEEE/CVF Conference on Computer Vision and Pattern Recognition}, pages 1921--1930, 2023.

\bibitem[Vaswani(2017)]{vaswani2017attention}
A Vaswani.
\newblock Attention is all you need.
\newblock \emph{Advances in Neural Information Processing Systems}, 2017.

\bibitem[Wang et~al.(2023{\natexlab{a}})Wang, Wang, Cao, Shen, and Huang]{wang2023painter}
Xinlong Wang, Wen Wang, Yue Cao, Chunhua Shen, and Tiejun Huang.
\newblock Images speak in images: A generalist painter for in-context visual learning.
\newblock In \emph{Proceedings of the IEEE/CVF Conference on Computer Vision and Pattern Recognition}, pages 6830--6839, 2023{\natexlab{a}}.

\bibitem[Wang et~al.(2023{\natexlab{b}})Wang, Zhang, Cao, Wang, Shen, and Huang]{wang2023seggpt}
Xinlong Wang, Xiaosong Zhang, Yue Cao, Wen Wang, Chunhua Shen, and Tiejun Huang.
\newblock Seggpt: Segmenting everything in context.
\newblock \emph{arXiv preprint arXiv:2304.03284}, 2023{\natexlab{b}}.

\bibitem[Wang et~al.(2024)Wang, Fang, Li, Li, Chen, and Liu]{wang2024skeleton}
Xinshun Wang, Zhongbin Fang, Xia Li, Xiangtai Li, Chen Chen, and Mengyuan Liu.
\newblock Skeleton-in-context: Unified skeleton sequence modeling with in-context learning.
\newblock In \emph{Proceedings of the IEEE/CVF Conference on Computer Vision and Pattern Recognition}, pages 2436--2446, 2024.

\bibitem[Wang et~al.(2023{\natexlab{c}})Wang, Jiang, Lu, He, Chen, Wang, Zhou, et~al.]{wang2023promptdiff}
Zhendong Wang, Yifan Jiang, Yadong Lu, Pengcheng He, Weizhu Chen, Zhangyang Wang, Mingyuan Zhou, et~al.
\newblock In-context learning unlocked for diffusion models.
\newblock \emph{Advances in Neural Information Processing Systems}, 36:\penalty0 8542--8562, 2023{\natexlab{c}}.

\bibitem[Wei et~al.(2022)Wei, Tay, Bommasani, Raffel, Zoph, Borgeaud, Yogatama, Bosma, Zhou, Metzler, et~al.]{wei2022emergent}
Jason Wei, Yi Tay, Rishi Bommasani, Colin Raffel, Barret Zoph, Sebastian Borgeaud, Dani Yogatama, Maarten Bosma, Denny Zhou, Donald Metzler, et~al.
\newblock Emergent abilities of large language models.
\newblock \emph{Transactions on Machine Learning Research}, 2022.

\bibitem[Xie and Tu(2015)]{xie2015hed}
Saining Xie and Zhuowen Tu.
\newblock Holistically-nested edge detection.
\newblock In \emph{Proceedings of the IEEE international conference on computer vision}, pages 1395--1403, 2015.

\bibitem[Xu et~al.(2023)Xu, Gandelsman, Bar, Yang, Gao, Darrell, and Wang]{xu2023improv}
Jiarui Xu, Yossi Gandelsman, Amir Bar, Jianwei Yang, Jianfeng Gao, Trevor Darrell, and Xiaolong Wang.
\newblock Improv: Inpainting-based multimodal prompting for computer vision tasks.
\newblock \emph{arXiv preprint arXiv:2312.01771}, 2023.

\bibitem[Yin et~al.(2024)Yin, Gharbi, Park, Zhang, Shechtman, Durand, and Freeman]{yin2024improved}
Tianwei Yin, Micha{\"e}l Gharbi, Taesung Park, Richard Zhang, Eli Shechtman, Fredo Durand, and William~T Freeman.
\newblock Improved distribution matching distillation for fast image synthesis.
\newblock \emph{arXiv preprint arXiv:2405.14867}, 2024.

\bibitem[Zhang et~al.(2018)Zhang, Isola, Efros, Shechtman, and Wang]{zhang2018lpips}
Richard Zhang, Phillip Isola, Alexei~A Efros, Eli Shechtman, and Oliver Wang.
\newblock The unreasonable effectiveness of deep features as a perceptual metric.
\newblock In \emph{CVPR}, 2018.

\bibitem[Zhang et~al.(2023)Zhang, Zhou, and Liu]{zhang2023vpr}
Yuanhan Zhang, Kaiyang Zhou, and Ziwei Liu.
\newblock What makes good examples for visual in-context learning?
\newblock \emph{Advances in Neural Information Processing Systems}, 36:\penalty0 17773--17794, 2023.

\bibitem[Zhou et~al.(2019)Zhou, Zhao, Puig, Xiao, Fidler, Barriuso, and Torralba]{zhou2019ade20k}
Bolei Zhou, Hang Zhao, Xavier Puig, Tete Xiao, Sanja Fidler, Adela Barriuso, and Antonio Torralba.
\newblock Semantic understanding of scenes through the ade20k dataset.
\newblock \emph{International Journal of Computer Vision}, 127:\penalty0 302--321, 2019.

\end{thebibliography}
}
\newpage

\setcounter{section}{0}
\def\thesection{\Alph{section}}

\section{Overview}

This document is structured as follows:
\begin{itemize}
    \item \Cref{supp_sec:related_work}: Related work
    \item \Cref{supp_sec:implementation}: Implementation details
    \item \Cref{supp_sec:add_quant}: Additional quantitative results
    \item \Cref{supp_sec:eval_painter}: Discussion on V-ICL models trained on task-related data
    \item \Cref{supp_sec:add_abl}: Additional ablations
    \item \Cref{supp_sec:limitations}: Limitations and future work
    \item \Cref{supp_sec:add_qual}: Additional qualitative results 
    
\end{itemize}

\section{Related Work}
\label{supp_sec:related_work}

In-context learning (ICL) has garnered significant attention in the field of natural language processing (NLP) with the advent of large-scale language models like GPT-3 \citep{brown2020gpt3} and its successors \citep{rae2021gopher, thoppilan2022lamda, chowdhery2023palm, touvron2023llama}. 
These models demonstrate the ability to perform tasks by conditioning on a small number of source-target examples, termed prompts, without any gradient updates or finetuning, effectively adapting to new tasks on-the-fly \citep{wei2022emergent, hao2022language}. 
The success of ICL in NLP has sparked interest in extending these capabilities to other domains, particularly in the realm of computer vision.

However, translating the concept of in-context learning from NLP to computer vision presents unique challenges due to the diversity in images and the inherent complexity of visual tasks.
This has led to the emergence of two primary schools of thought in adapting ICL to computer vision, termed visual in-context learning (V-ICL).

The first approach adapts vision foundation models for in-context learning by training on uncurated datasets composed of random crops that potentially include examples of source images and corresponding targets (\eg figures from computer vision papers). 
Research such as Visual Prompting \citep{bar2022visualprompting} and IMProv \citep{xu2023improv} exemplify this approach, where they train a ViT-based MAE-VQGAN architecture \citep{he2022mae, esser2021vqgan} on the task of masked inpainting. 
During inference, these methods involve creating composite images by stitching together a query image with prompt examples, forming a grid-like structure with a placeholder mask for the prediction, that the inpainting model can process. 
While these methods yield promising results, this approach often suffers from weaker inference of context between the query image and the prompt, lower resolution predictions, and overall weaker prediction quality.

The second school of thought aims to enhance prediction performance by training vision foundation models on curated/annotated task-related datasets. 
This method involves training/finetuning a model but uses paired source-target images of multiple tasks as training data. 
Notable examples of this method include Painter \citep{wang2023painter}, Prompt Diffusion \citep{wang2023promptdiff}, SegGPT \citep{wang2023seggpt}, Skeleton-In-Context \citep{wang2024skeleton}, and Point-In-Context \citep{fang2024point}. 
While models such as Painter and Prompt Diffusion target a relatively diverse set of tasks, the others focus on building generalist models to cater specific tasks such as segmentation, skeleton sequence modeling, or 3D point cloud estimation. 
Although these models achieve improved results and provide important insights for future research in visual in-context learning, they require updating model weights using datasets related to the out-of-domain tasks.
This in turn implies the need for training data on related out-of-domain tasks that we are trying to adapt to. 
We believe that this ideology diverges from the core principles of ICL as they often fall short in generalizing to novel tasks that are unrelated to the training set and rely on large annotated datasets. 
This approach, therefore, somewhat undermines the fundamental idea of ICL, which emphasizes the ability to adapt to new tasks without retraining nor requiring a large annotated dataset.

\section{Implementation Details} \label{supp_sec:implementation}

\textbf{SD-VICL:} 
We base our experiments on an off-the-shelf Stable Diffusion model \citep{rombach2022stablediff}, specifically the v1.5 checkpoint.
Unless specified otherwise, we use the following hyperparameters for all our evaluations: denoising time steps ($T$) = 70, attention temperature ($\tau$) = 0.4, contrast strength ($\beta$) = 1.67, and swap-guidance scale ($\gamma$) = 3.5. 
Further, we set the text condition of the Stable Diffusion pipeline to an empty string, and thus, no supplementary guidance is provided beyond the input prompts. 

\vspace{0.4em}\noindent\textbf{Comparison baselines:}
We use the publicly available repositories and checkpoints for Visual Prompting \citep{bar2022visualprompting}, IMProv \citep{xu2023improv}, Painter \citep{wang2023painter}, LVM \citep{bai2024lvm}, and Prompt Diffusion \citep{wang2023promptdiff} to generate the results for all the experiments. 
For the text-guided variant of IMProv, as specified in their paper, we provide the model with a string comprising of the location and task information (\eg ``Left - input image, right
- Black and white foreground/background segmentation").
To ensure a fair comparison, all methods, including ours, are evaluated using the same set of prompts, which we obtain using the unsupervised prompt retrieval method outlined by \citet{zhang2023vpr}.

\vspace{0.4em}\noindent\textbf{Tasks and datasets:}
Below, we provide details on the tasks and datasets used for evaluations in our experiments:

\begin{itemize}
    \item \textbf{Foreground segmentation:} 
    This is a binary segmentation task, which predicts a binary mask of the object of interest (\ie foreground) in an image. 
    The prompt groundtruth is a black-and-white image with the foreground being white and the background being black. 
    For evaluation, we use the Pascal-5i dataset \citep{shaban2017pascal5i}, which comprises of 1864 images belonging to 20 object classes. 
    The images are divided into four splits, where each split consists of five unique classes. 
    We use the \textit{mean intersection-over-union} (mIoU) as the evaluation metric. 
    
    \item \textbf{Single object detection:} 
    This task is similar to the foreground segmentation task, however, in this task, the bounding box of the object of interest is predicted instead of the mask with the exact boundary. 
    For this task, the prompt groundtruth is a black-and-white image with the bounding box colored in white. 
    We use the same dataset as foreground segmentation but include only images with single instances of objects following \citep{bar2022visualprompting, xu2023improv}. 
    The subset thus chosen consists of 1312 images and we report the mIoU scores.

    \item \textbf{Semantic segmentation:}
    This task predicts the per-pixel semantic label of a given image. 
    We follow the method proposed by \citet{wang2023painter} to compose the prompt groundtruth, which assigns equally-spaced unique colors to each class. 
    We use the Cityscapes dataset \citep{cordts2016cityscapes}, which consists of 19 classes (excluding the void classes), and the COCOStuff dataset \citep{lin2014coco}, which consists of 27 mid-level classes. 
    We report the mIoU and pixel accuracy scores as evaluation metrics.

    \item \textbf{Keypoint detection:}
    The task of keypoint detection entails locating the critical points or landmarks of an object. 
    In this study, we focus on human pose keypoint detection, which predicts the locations of the 17 keypoints defined in COCO \citep{lin2014coco}.
    Since the prompt groundtruth needs to be in the form of an image, we create an image that depicts the keypoints in the form of a heatmap as shown in Fig. 4. 
    Each heatmap is created by superimposing Gaussian distributions centered on each keypoint.
    To accommodate the different spatial scales, we apply Gaussians with smaller variance for facial keypoints, which are relatively finer, and larger variance for body keypoints. 
    These are visualized in two color channels: red for facial keypoints and green for body keypoints, facilitating easier decoding.
    For evaluation, following \citet{hedlin2024stablekeypoints}, we use the DeepFashion dataset \citep{liu2016deepfashion} and report metrics: \textit{mean sqaured error} (MSE) and the \textit{percentage of correct keypoints} (PCK). 

    \item \textbf{Edge detection:}
    The goal of this task is to predict the boundaries and edges within an image. 
    For evaluation, we utilize the validation set of the NYUDv2 dataset \citep{silberman2012nyudv2} comprising 654 images. 
    Since the validation set did not have the groundtruth, we used the soft edge maps generated using HED \citep{xie2015hed} as the pseudo-groundtruth. 
    For evaluation we compute the MSE and the LPIPS loss \citep{zhang2018lpips} between the HED-predicted edge map and the V-ICL predictions.

    \item \textbf{Colorization:} 
    In this task, the objective is to colorize a given grayscale image.
    Similar to \citep{bar2022visualprompting, xu2023improv} we randomly sample 1000 images from the validation set of ImageNet \citep{russakovsky2015imagenet} for evaluation. 
    We compute the LPIPS loss and the FID score \citep{heusel2017fid} between the original colored image and the colorized prediction to evaluate the perceptual similarity.

\end{itemize}

\begin{table*}[!t]
  \centering
\resizebox{0.9\linewidth}{!}{

  \begin{tabular}{l|cccc|c|cccc|c}
    \toprule    
    Model 
    & \multicolumn{5}{c}{Foreground Segmentation} (mIoU~$\uparrow$) 
    & \multicolumn{5}{c}{Single Object Detection (mIoU~$\uparrow$)} \\

    & Split 0 & Split 1 & Split 2 & Split 3 & Avg. 
    & Split 0 & Split 1 & Split 2 & Split 3 & Avg. \\
    
    \midrule
    
    \multicolumn{4}{l}{Number of Example Prompts: 1}  \vspace{2pt} \\
    
    \quad Visual Prompting \citep{bar2022visualprompting}
    & 34.85 & 38.55 & 34.51 & 32.24 & 35.04  
    & \underline{48.82} & \underline{48.52} & \underline{45.11} & \underline{42.72} & \underline{46.29} \\
                         
    \quad IMProv (w/o text) \citep{xu2023improv}
    & \underline{41.46} & 43.60 & 39.70 & 33.22  & 39.50
    & 46.10 & 47.26 & 41.97 & 39.96 & 43.82 \\
    
    \quad IMProv (w/ text) \citep{xu2023improv}
    & 41.31 & \underline{44.64} & \underline{40.86} & \underline{35.93} & \underline{40.69}
    & 44.69 & 48.10 & 44.53 & 40.34 & 44.42 \\
    
    \quad SD-VICL (ours) 
    & \textbf{44.05} & \textbf{45.17} & \textbf{44.36} & \textbf{42.11} & \textbf{43.92}
    & \textbf{54.45} & \textbf{52.92} & \textbf{51.56} & \textbf{47.27} & \textbf{51.55} \\

    \midrule
    
    \multicolumn{8}{l}{Number of Example Prompts: 5} \vspace{2pt} \\
    
    \quad Visual Prompting \citep{bar2022visualprompting}
    & 36.70 & 40.02 & 36.18 & 32.56 & 36.37 
    & 51.59 & 49.30 & 46.80 & 44.66 & 48.09 \\

    \quad SD-VICL (ours)
    & \textbf{55.55} & \textbf{56.08} & \textbf{55.84} & \textbf{54.49} & \textbf{55.49}
    & \textbf{58.99} & \textbf{56.31} & \textbf{57.09} & \textbf{56.01} & \textbf{57.10}\\
 
    \bottomrule
  \end{tabular}
  }
  \caption{Quantitative performance comparison of the proposed approach with recent approaches on foreground segmentation and single object detection for each split of the Pascal-5i dataset. 
  }
  \label{supp_tab:quant_splits}
\end{table*}

\section{Additional Quantitative Results} \label{supp_sec:add_quant}
While in Tab. 1 we present the average performance for foreground segmentation and single object detection across all splits of the Pascal-5i dataset, in \cref{supp_tab:quant_splits}, we report the metrics for each split. 

Additionally, in Tab. 1, we present the results of single object detection evaluated on the entire dataset for a more generalized assessment. 
However, in \cref{supp_tab:objdet_50}, we follow the approach of \citet{bar2022visualprompting} and evaluate single object detection on a subset of the Pascal-5i dataset, where images with objects covering more than 50\% of the image are excluded. 
While we observe an overall drop in absolute scores for all methods, the performance trends remain consistent with Tab. 1. 
This decline in performance can be attributed to the fact that larger objects are generally easier to detect than smaller ones, as noted by \citet{bar2022visualprompting} as well.

Furthermore, we evaluated semantic segmentation on the COCOStuff dataset \citep{lin2014coco}, where we report the results in \cref{supp_tab:semseg_coco}. 
In contrast to the trend observed in Tab. 2, where performance improved with five example prompts compared to the single prompt, we could see a performance deterioration with five prompts in this case. 
Upon analysis, we identified that this performance decline was caused by the inconsistencies in labeling within the dataset, which creates confusion when inferring the context with multiple prompts, thereby negatively impacting the results.

\begin{table}[t]
  \centering
\resizebox{0.95\linewidth}{!}{

  \begin{tabular}{l|cccc}
    \toprule    
    Model &  \multicolumn{4}{c}{Single Object Detection (mIoU~$\uparrow$)} \\
         & Split 0 & Split 1 & Split 2 & Split 3  \\
    \midrule
    \multicolumn{4}{l}{Number of Example Prompts: 1}  \vspace{2pt} \\

    \quad Visual Prompting & \underline{42.94} & \underline{35.02} & \underline{37.77} & \underline{32.76}  \\
                         
    \quad IMProv (w/o text) & 42.32 & 36.52 & 36.32 & 31.83  \\
    
    \quad IMProv (w/ text) & 40.61 & 35.79 & 38.74 & 32.55 \\
    
    \quad Ours & \textbf{47.74} & \textbf{39.86} & \textbf{44.93} & \textbf{37.92} \\

    \midrule
    \multicolumn{4}{l}{Number of Example Prompts: 5} \vspace{2pt} \\
    
    \quad Visual Prompting & 45.07 & 34.86 & 38.37 & 34.23 \\

    \quad Ours & \textbf{51.74} & \textbf{43.15} & \textbf{50.23} & \textbf{47.20} \\
    
    \bottomrule
  \end{tabular}
  }
  \caption{Quantitative evaluation of single object detection on a subset of the Pascal-5i dataset, where larger objects with an area greater than 50\% were excluded.}
  \label{supp_tab:objdet_50}
\end{table}

\section{Discussion on V-ICL Models Trained on Task-Related Data}\label{supp_sec:eval_painter}

As emphasized in the main paper, our work the first to propose a fully training-free paradigm that \textit{uncovers} the V-ICL properties of a vision foundation model. 
For fairness, Sec. 3.1 of the main paper evaluates our approach against Visual Prompting \citep{bar2022visualprompting} and IMProv \citep{xu2023improv}, as they are the closest in methodology.  
While these models involve training, they do so on uncurated datasets, unlike models such as Painter \citep{wang2023painter}, LVM \citep{bai2024lvm}, and Prompt Diffusion \citep{wang2023promptdiff}, which are trained on task-related annotated data.

To ensure completeness, we extend our evaluations to these V-ICL models trained on task-related data. 
Painter leverages a ViT-Large \citep{dosovitskiy2020vit} backbone trained on multiple annotated datasets (\eg COCO \citep{lin2014coco}, ADE20K \citep{zhou2019ade20k}, and NYUv2 \citep{silberman2012nyudv2}). 
LVM is built on OpenLLaMA's 7B model \citep{openlm2023openllama} and trained on the UVD-V1 \citep{bai2024lvm} dataset, a large-scale vision corpus comprising 50 datasets (\eg LAION5B \citep{schuhmann2022laion}) that span annotated, unannotated, and sequence images. 
Prompt Diffusion is a generative model based on Stable Diffusion, jointly finetuned on three forward tasks (\ie image-to-depth, image-to-edge, image-to-segmentation) and their inverse variants, using vision-language prompts with paired images and text guidance. 
The training is conducted on a dataset adapted from \citet{brooks2023instructpix2pix}.

\begin{table}[!t]
  
  \centering
\resizebox{0.8\linewidth}{!}{

  \begin{tabular}{l|cc}
    \toprule    
    Model &  \multicolumn{2}{c}{Semantic Segmentation} \\
         & mIoU~$\uparrow$ & Acc.~$\uparrow$ \\
    \midrule

    \multicolumn{2}{l}{Number of Example Prompts: 1}  \vspace{2pt} \\

    \quad Visual Prompting & 15.31 & 39.07 \\
                         
    \quad IMProv (w/o text) & 17.09 & 41.64 \\
    
    \quad IMProv (w/ text) & \underline{17.19} & \underline{42.35} \\
    
    \quad Ours & \textbf{28.32} & \textbf{56.84} \\

    \midrule

    \multicolumn{2}{l}{Number of Example Prompts: 5} \vspace{2pt} \\
    
    \quad Visual Prompting & 13.01 & 36.12 \\
                           
    \quad Ours &  \textbf{21.80} & \textbf{53.01} \\

    \bottomrule
  \end{tabular}
  }
  \caption{Quantitative evaluation of semantic segmentation on the COCOStuff dataset.}
  \label{supp_tab:semseg_coco}
\end{table}

\begin{figure*}[!t]
    \centering
    \includegraphics[width=\linewidth]{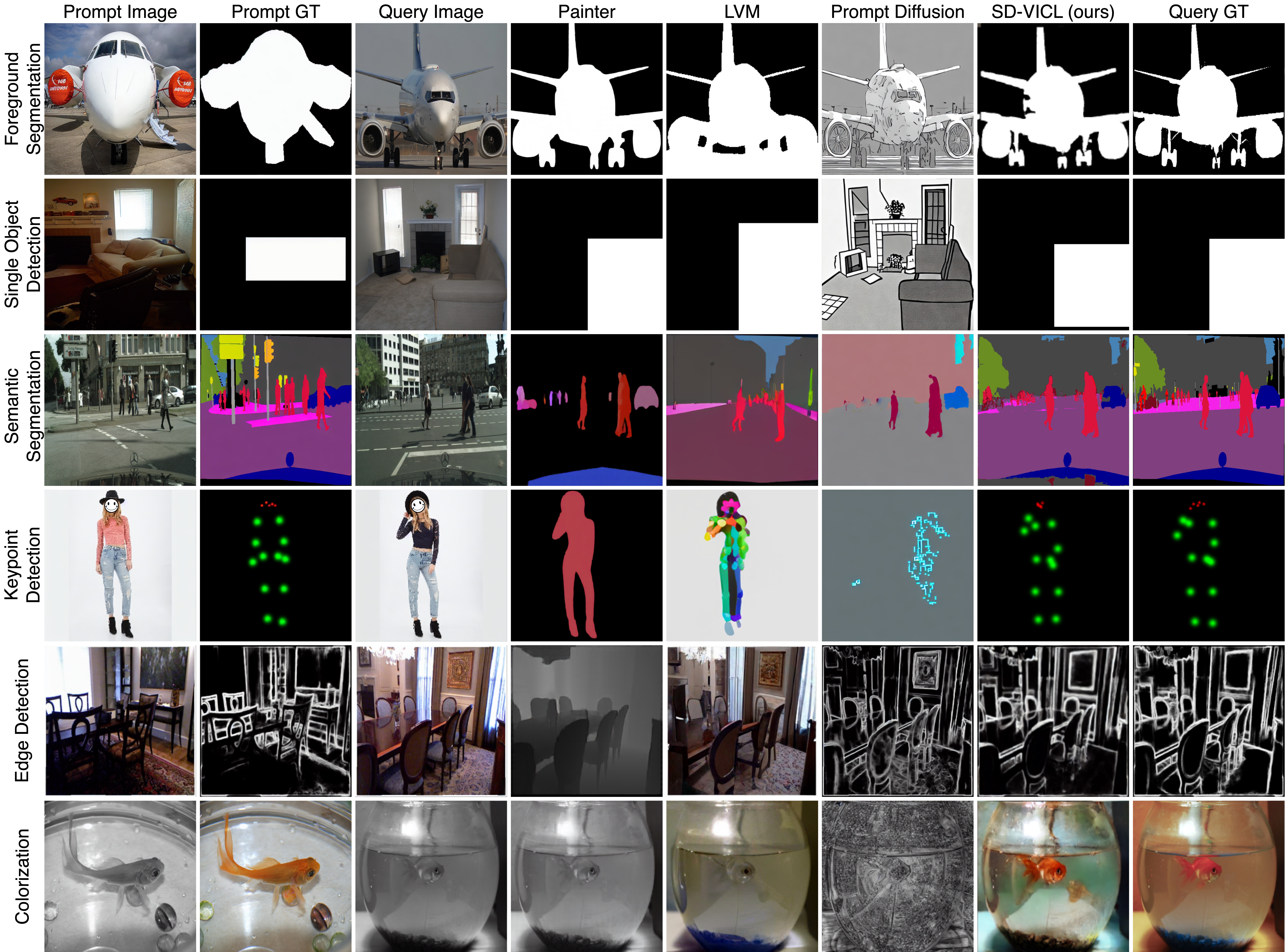}
    \caption{Additional qualitative comparisons illustrating the performance of training-based V-ICL models, Painter \citep{wang2023painter}, LVM \citep{bai2024lvm}, and Prompt Diffusion \citep{wang2023promptdiff} our proposed method on six different tasks. It can be seen that our method produces visually superior results as compared to  the baselines.}
    \vspace{-10pt}
    \label{supp_fig:qualitative_painter_lvm}
\end{figure*}

The quantitative and qualitative comparisons are presented in Tab. 2 and \cref{supp_fig:qualitative_painter_lvm}, respectively. 
Overall, we observe that our method outperforms all three baselines across multiple tasks. 

We observe that all three modes often suffer from overfitting to training tasks, leading to poor generalization when exposed to novel tasks.
Although visual in-context learning should ideally infer the task from the relationship between the prompt image and its groundtruth, these models demonstrate weakness in this regard. 

\begin{figure*}[!t]
    \centering
    \begin{subfigure}[b]{0.9\linewidth}
    \includegraphics[width=\linewidth]{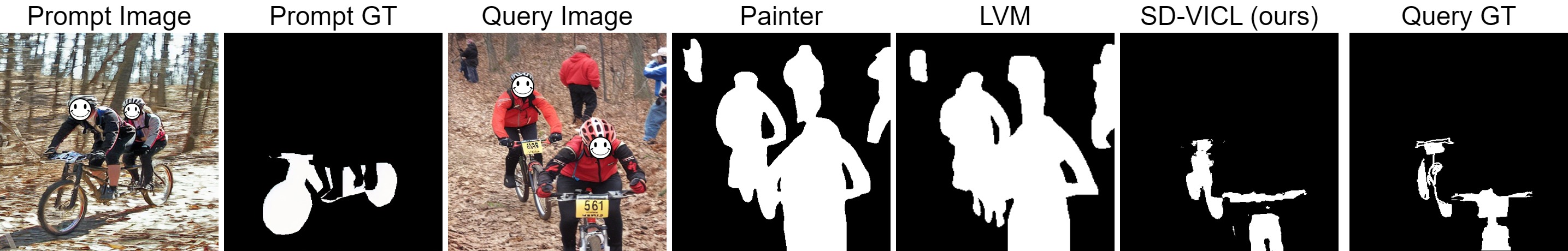}
        \caption{\phantom{(a)}}
        \label{supp_fig:painter_lvm_fail_set1}
    \end{subfigure}
    \begin{subfigure}[b]{0.95\linewidth}
        \includegraphics[width=\linewidth]{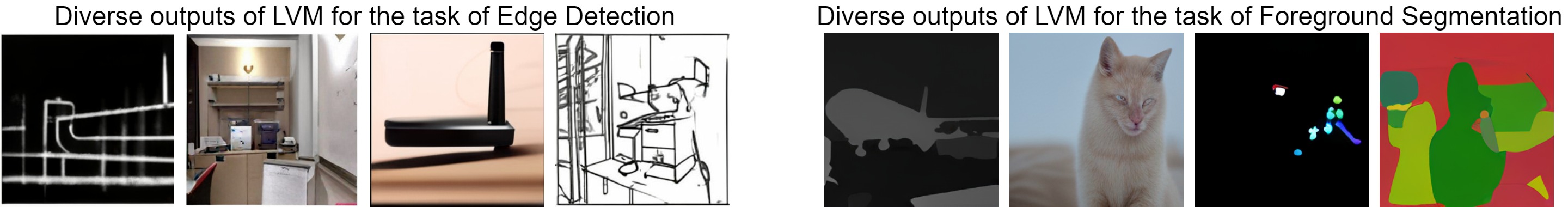}
        \caption{\phantom{(b)}}
        \label{supp_fig:painter_lvm_fail_set2}
    \end{subfigure}
    \caption{\textbf{Failure cases of V-ICL models trained on task-related data}, Painter \citep{wang2023painter} and LVM \citep{bai2024lvm}, implying poor task inference. 
    In (a), both models fail in multi-class scenarios, segmenting the entire foreground instead of focusing on the region of interest defined by the prompt image and its corresponding ground truth. 
    Examples in (b), depict inconsistent outputs generated by LVM for the same task (left: edge detection, right: foreground segmentation). The inputs for each of these outputs adhered to the same format as shown in \cref{supp_fig:qualitative_painter_lvm}, yet LVM produces outputs in diverse domains, deviating from the domain of the prompt groundtruth. These cases further emphasize the poor task inference capabilities of Painter and LVM.}
    \label{supp_fig:fail_painter_lvm}
    \vspace{-10pt}
\end{figure*}

Painter performs well on simple tasks like foreground segmentation and object detection when the query image contains a single foreground category (\cref{supp_fig:qualitative_painter_lvm}, row 1). 
However, in multi-class scenarios (\cref{supp_fig:painter_lvm_fail_set1}), Painter segments the entire foreground rather than focusing on the specific region of interest defined by the relationship between the prompt image and its groundtruth.
Further, overfitting to training tasks is evident in rows 3 and 4 of \cref{supp_fig:qualitative_painter_lvm}, where Painter outputs a segmentation map in semantic segmentation with a different color scheme than defined in the prompt groundtruth. 
Similarly, for keypoint detection, Painter outputs a segmentation map instead of a heatmap for keypoints.
Moreover, Painter struggles with colorization, often outputting the grayscale image itself. 
In edge detection, Painter outputs a depth map instead of the expected edge map (\cref{supp_fig:qualitative_painter_lvm}, row 2). 
This behavior suggests overfitting to the NYUv2 dataset, where the edge map query/prompt images overlap with those used for depth estimation during their training.

Similar limitations are observed for LVM, including poor performance on multi-class foreground segmentation (\cref{supp_fig:painter_lvm_fail_set1}), overfitting to training tasks (\cref{supp_fig:qualitative_painter_lvm}, row 4), and lack of generalization. 
Additionally, LVM exhibits inconsistencies in its outputs, as shown in \cref{supp_fig:painter_lvm_fail_set2}. 
Specifically, for a given task, despite the format/domain of the inputs remaining unchanged, we observe that the generated outputs belong to diverse domains. 
For example, in foreground segmentation, while some outputs align with foreground segmentation, others unexpectedly belong to unrelated domains such as keypoints, segmentation maps, or RGB images. 
This inconsistency highlights LVM's inability to produce coherent predictions despite the task and input format remaining unchanged.

Prompt Diffusion, while aiming to unlock in-context capabilities via vision-language prompts, remains constrained by the six tasks it is explicitly trained on. 
Although it exhibits relatively stronger performance in edge detection and segmentation --- tasks included in its training, it struggles on tasks outside this scope.  
For instance, Prompt Diffusion produces structurally incoherent outputs for keypoint detection, object detection, and colorization, failing to align with the semantics illustrated by the prompt pair.  
It also occasionally hallucinates incorrect colors, textures, or layouts, especially when confronted with novel task types or subtle prompt-query domain shifts.

These observations highlight shared limitations among Painter, LVM, and Prompt Diffusion in inferring tasks and context purely from input prompts. 
Their reliance on task-specific training data results in overfitting, leading to poor generalization on novel tasks. 
In contrast, our proposed training-free method demonstrates robust generalization and effective task inference, underscoring the benefits of uncovering V-ICL properties without additional training and the superiority of the proposed method to explicitly infer the context and task from the inputs, as intended by V-ICL.

\section{Additional Ablations} \label{supp_sec:add_abl}

In addition to the ablations discussed in the main paper, we also experimented with alternative attention formulations and the effects of several other factors such as temperature hyperparameter, resolution of the self-attention layers, contrastive strength parameter, swap-guidance scale, and AdaIN.

\begin{figure}[t!]
    \centering
    \includegraphics[width=\linewidth]{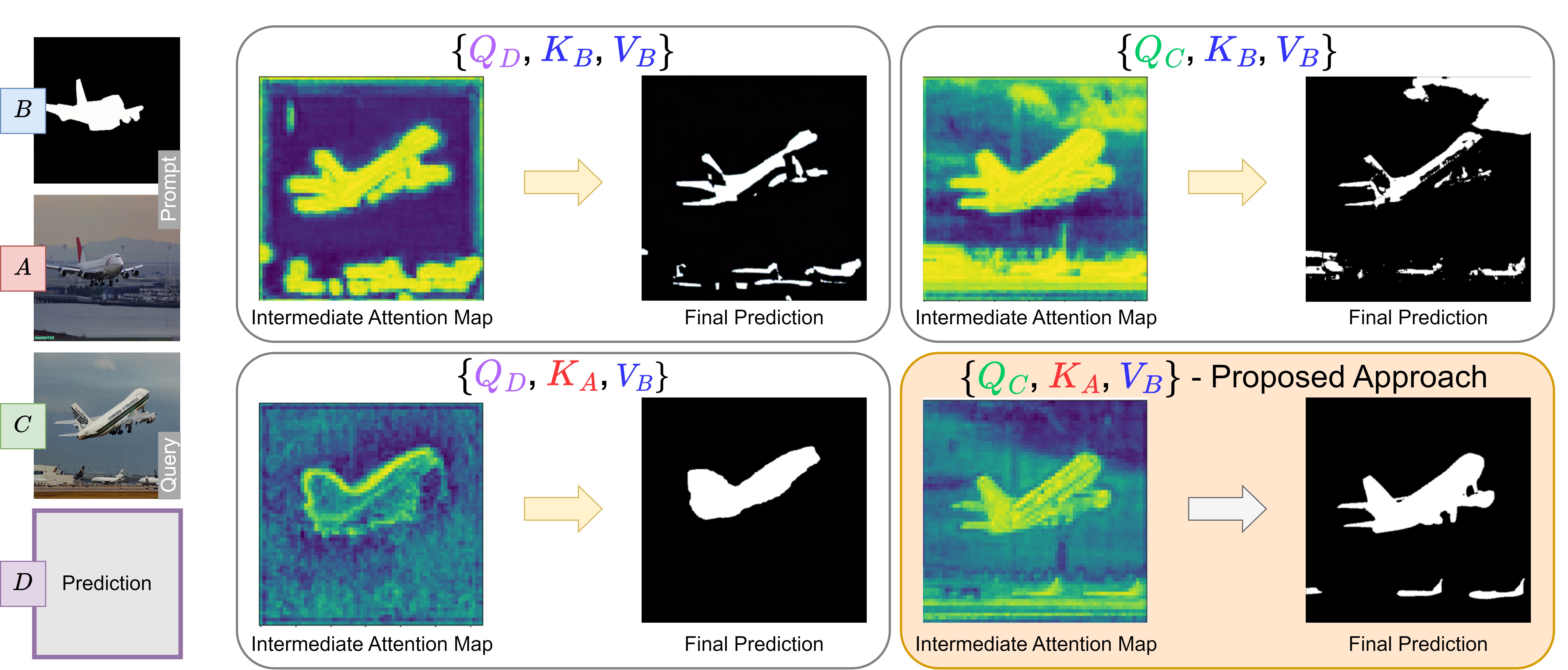}
    \caption{Qualitative examples of alternative attention formulations.
    }\vspace{-15pt}
    \label{supp_fig:abl_attn}
\end{figure}

\vspace{0.4em}\noindent\textbf{Alternative attention formulations:} With regards to the attention formulation between query and the prompt, there are potentially multiple variants that could be used instead of the one described by Eq. (7). 
These candidate formulations can be derived by substituting the $Q$ and $K$ of Eq. (7) with the corresponding elements of each of these sets: $\{Q_D, K_B\}$, $\{Q_C, K_B\}$, and $\{Q_D, K_A\}$. 
Since the prediction needs to correspond to the features of the prompt groundtruth, the value vector, $V$, needs to come from $B$ and cannot be substituted with other alternate options. 
\cref{supp_fig:abl_attn} illustrates a subset of these variants along with the predictions obtained using these alternate formulations. 
The quantitative performance corresponding to these candidate formulations are presented in  \cref{supp_tab:abl_attn}.  
However, since the prompt groundtruth lacks semantics, such formulations (\ie $\{Q_D, K_B, V_B\}$, $\{Q_C, K_B, V_B\}$) tend to focus on color similarities rather than inferring the underlying semantic correlations. 
Alternatively, we can formulate the attention using the Query vector from the prediction ($D$) itself, similar to the approach followed by \citet{alaluf2024cross}.
In this scenario, the intermediate predictions at early denoising stages closely resemble those produced by our formulation.
However, in later denoising stages, the performance deteriorates as the prediction gradually shifts towards the prompt groundtruth, which lacks semantics, impairing the prediction performance. 
As seen in \cref{supp_fig:abl_attn} and \cref{supp_tab:abl_attn}, the proposed formulation demonstrates superior performance which is achieved by ensuring that at each denoising step, the process is guided by the query and prompt latents at the corresponding denoising stages, thereby preserving the essential semantics needed for better context and task inference. 

\begin{table}[!t]
\centering
\resizebox{0.6\linewidth}{!}{
\begin{tabular}{c|c}
         \toprule
         Method & mIoU~$\uparrow$ \\
         \midrule
         $\{Q_C, K_B, V_B\}$ & 12.54 \\
         $\{Q_D, K_B, V_B\}$ & 12.95 \\
         $\{Q_D, K_A, V_B\}$ & 23.68 \\
         Proposed: $\{Q_C, K_A, V_B\}$ & \textbf{55.49} \\
         \bottomrule
\end{tabular}
}
\caption{Ablation of attention formulations on foreground segmentation evaluated on Pascal-5i.}\label{supp_tab:abl_attn}
\end{table}

\begin{figure*}[!t]
    \centering
    \scalebox{0.85}{
    \begin{subfigure}{0.32\linewidth}
        \centering
        \includegraphics[width=\linewidth]{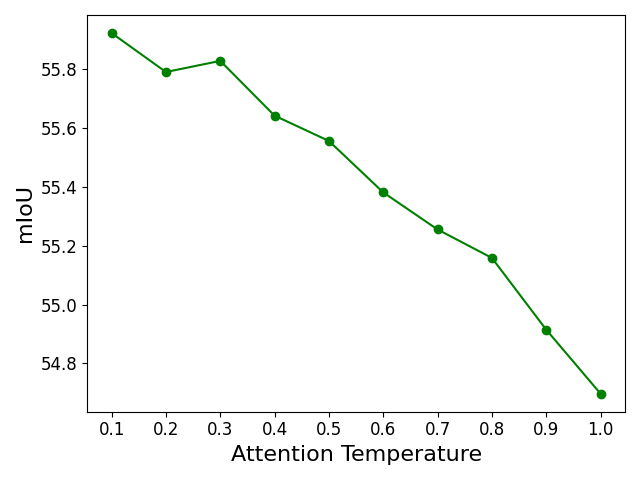}
        \caption{}
    \end{subfigure}
    \hfill
    \begin{subfigure}{0.32\linewidth}
        \centering
        \includegraphics[width=\linewidth]{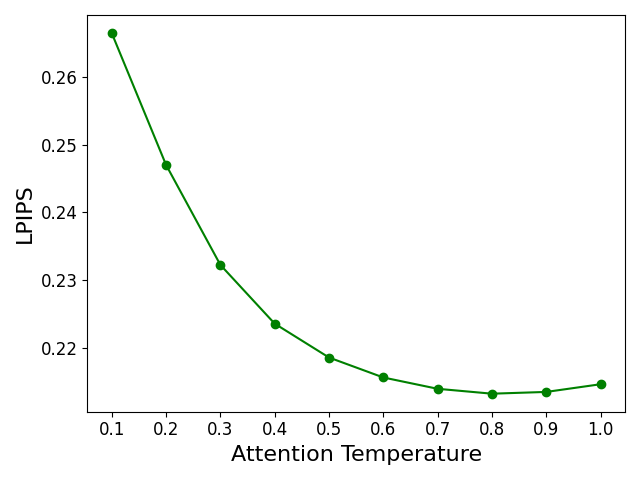}
        \caption{}
    \end{subfigure}
    \hfill
    \begin{subfigure}{0.32\linewidth}
        \centering
        \includegraphics[width=\linewidth]{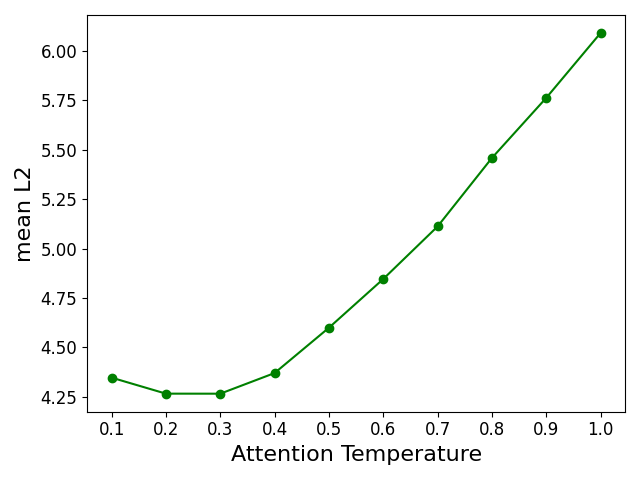}
        \caption{}
        
    \end{subfigure}
    }
    \caption{We illustrate the performance variation with respect to the attention temperature hyperparameter for the following tasks: \\(a) foreground segmentation, (b) colorization, and (c) keypoint detection. \label{supp_fig:temp_para}
    }
\end{figure*}

\vspace{0.4em}\noindent\textbf{Temperature hyperparameter, $\tau$:}
As shown in Eq. (7), we introduce a temperature hyperparameter ($\tau$) to the attention computation in order to control the sharpness of correspondence between the patches of the query image and the prompt image. 
While we use a constant temperature hyperparameter (\ie $\tau = 0.4$) across all tasks to preserve generalization, we investigated the effect of $\tau$ on the performance of a few proxy tasks.
We observe that the optimal temperature parameter varies notably with the task, which we depict in \cref{supp_fig:temp_para}.

\begin{figure}[t]
    \centering
    \begin{subfigure}{0.49\linewidth}
        \centering
        \includegraphics[width=\linewidth]{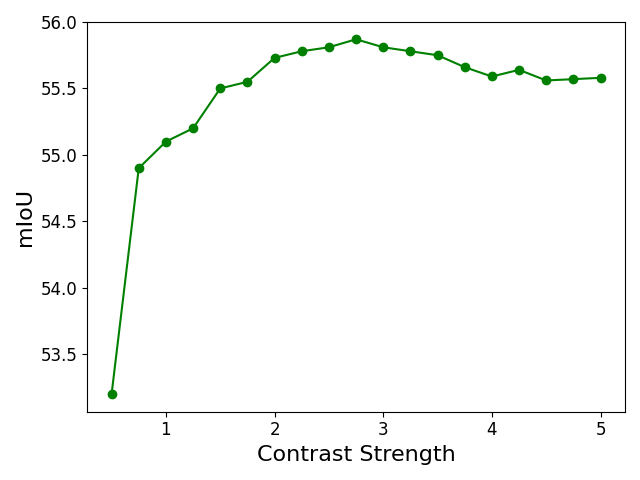}
        \caption{}
    \end{subfigure}
    \hfill
    \begin{subfigure}{0.49\linewidth}
        \centering
        \includegraphics[width=\linewidth]{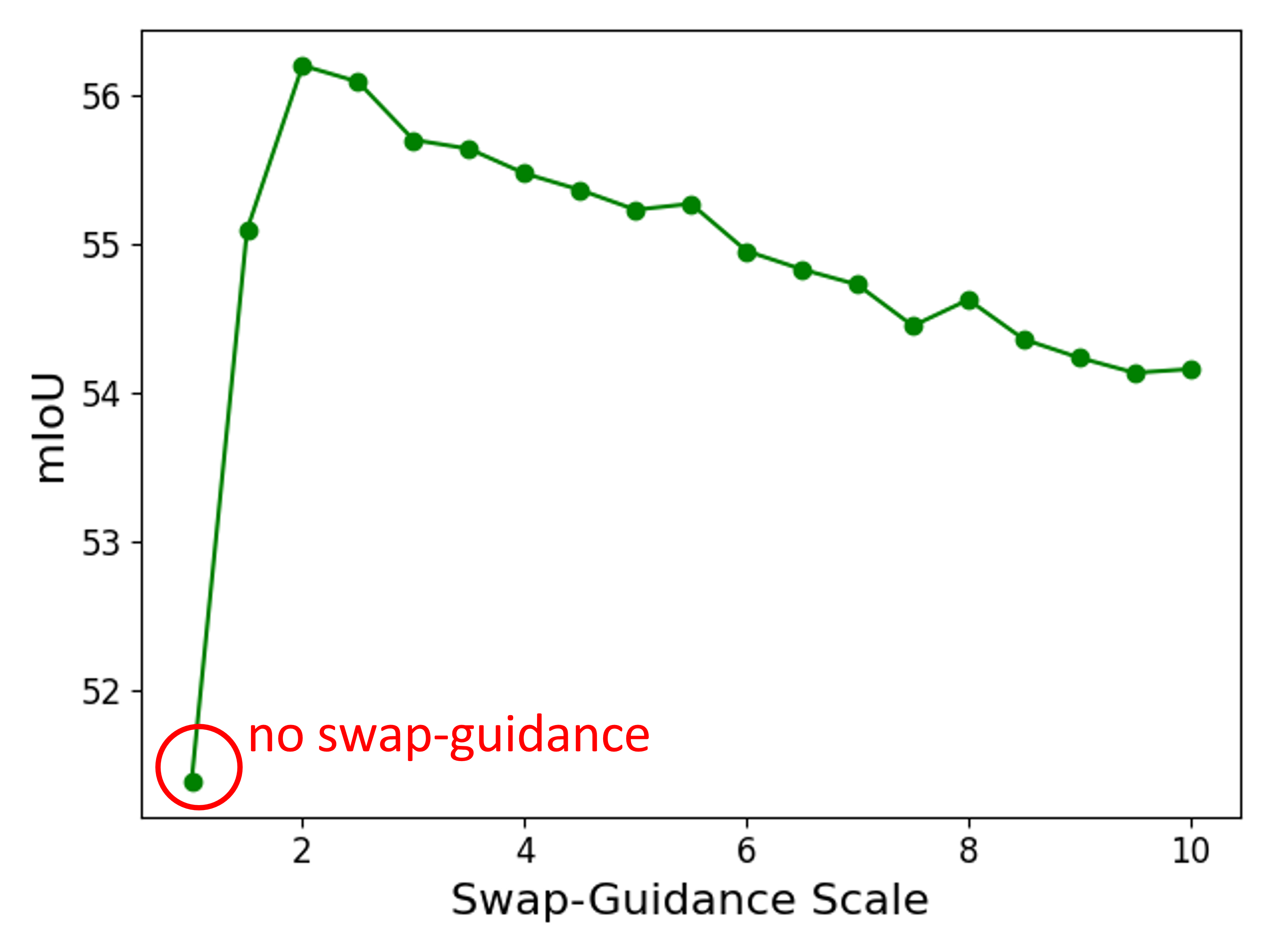}
        \caption{}
    \end{subfigure}
    \caption{Performance variation with respect to (a) contrast strength and (b) swap-guidance scale hyperparameters.  \label{supp_fig:cont_swap}
    }
\end{figure}

\vspace{0.4em}\noindent\textbf{Contrast strength ($\beta$) and swap-guidance scale ($\gamma$) hyperparameters:}
We adapt the \textit{attention map contrasting} (Eq. (8)) and \textit{swap-guidance} (Eq. (9)) methods from \citet{alaluf2024cross} to address the domain gap introduced by using multiple images from different domains (\ie source and target images belong to distinct domains). 
While we utilize the hyperparameter values proposed by \citep{alaluf2024cross} (\ie $\beta = 1.67 , \gamma = 3.5 $), we investigate their impact on performance using foreground segmentation as a proxy task.
We depict the variation of the performance with respect to the contrast strength and the swap-guidance scale in \cref{supp_fig:cont_swap}. 
A notable improvement in performance can be observed with a contrast strength greater than 1.0 and with swap-guidance enabled. 

\begin{figure}[!t]
    \centering
    \includegraphics[width=0.7\linewidth]{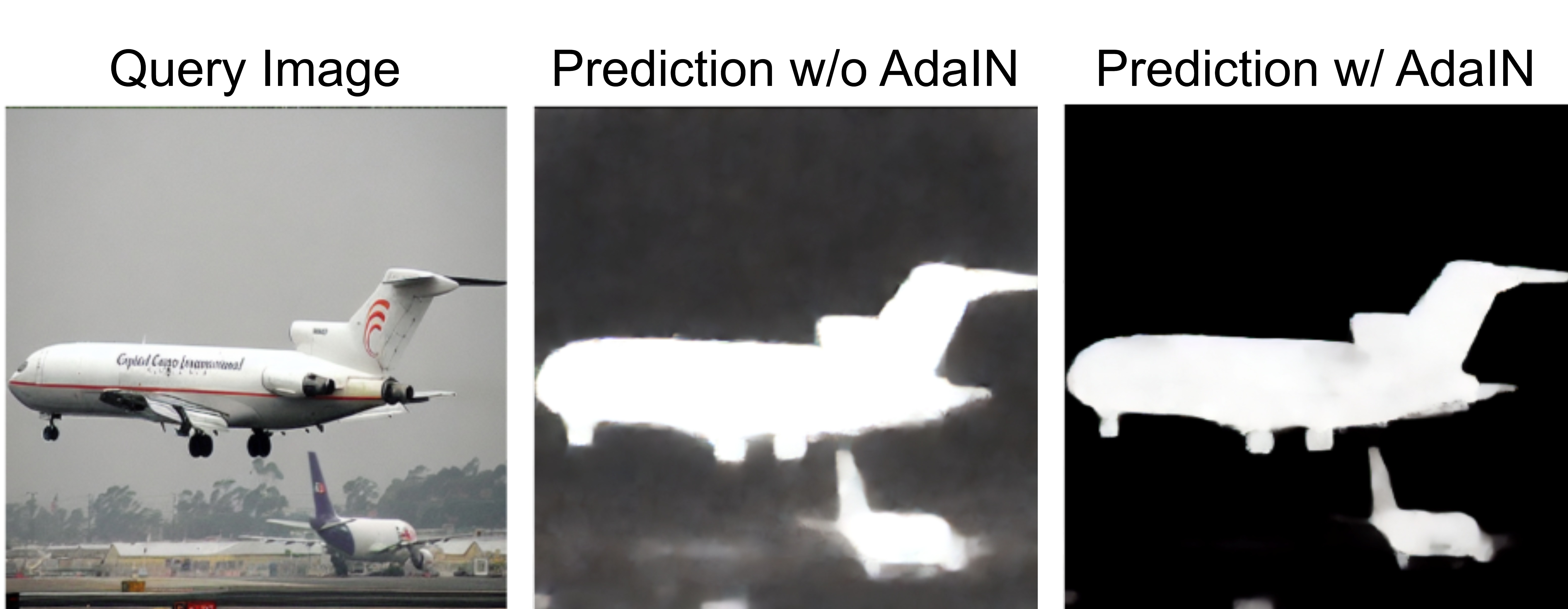}  
    \caption{Example comparing the prediction with and without AdaIN.} \label{supp_fig:adain}
\end{figure}

\vspace{0.4em}\noindent\textbf{Adaptive instance normalization (AdaIN):}
As explained in Sec. 2.2, we utilize AdaIN to align the color distribution between the prediction ($D$), which is initialized using the noise space of the query image ($C$), and the expected groundtruth color space (\ie color space of $B$). 
In \cref{supp_fig:adain} we present a comparison example with and without AdaIN, and in \cref{supp_tab:adain} we tabulate the overall performance on foreground segmentation. 
A clear performance improvement could be observed with the incorporation of AdaIN. 

\begin{table}[!t]
\centering
    \resizebox{0.4\linewidth}{!}{
    \begin{tabular}{c|cc}
        \toprule
        Model & mIoU~$\uparrow$ \\
        \midrule
        w/o AdaIN & 51.55 \\
        w/ AdaIN & 55.49 \\
        \bottomrule
    \end{tabular}
    }
    \caption{Quantitative evaluation of with and without AdaIN evaluated using foreground segmentation.} \label{supp_tab:adain}
\end{table}

\begin{table}[!t]
  
  \centering
\resizebox{0.7\linewidth}{!}{

  \begin{tabular}{ccc|c}
    \toprule 
    \multicolumn{3}{c}{Resolution} & mIoU~$\uparrow$ \\
    $16\times 16$ & $32 \times 32$ & $64 \times 64$ & \\
    \midrule
    \checkmark & - & - & 11.39 \\
    - & \checkmark & - & 32.50 \\
    - & - & \checkmark & 50.33 \\
    \checkmark & \checkmark & - & 35.48 \\
    \checkmark & - & \checkmark &  52.52 \\
    - & \checkmark & \checkmark & 53.76 \\
    \checkmark & \checkmark & \checkmark & \textbf{55.49} \\
    
    \bottomrule
  \end{tabular}
  }
  \caption{Quantitative evaluation on different combinations of resolutions which the self-attention layers could be modified using the proposed in-place attention reformulation.}
  \label{supp_tab:attn_layers}
\end{table}

\begin{figure*}[t!]
    \centering
    \includegraphics[width=0.97\linewidth]{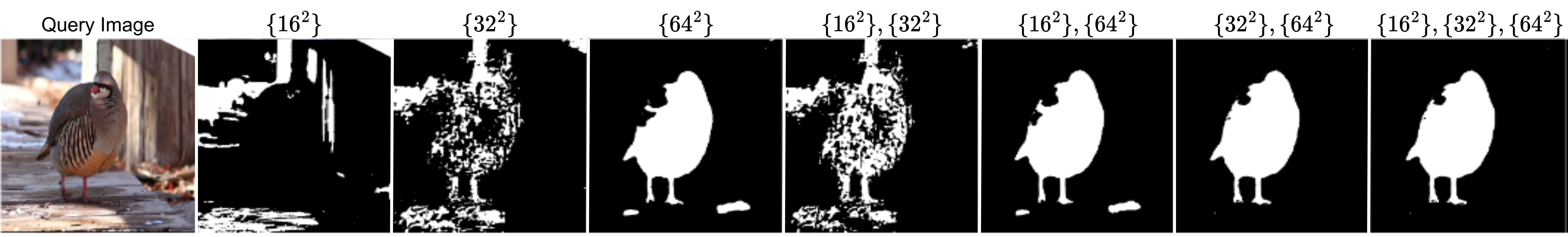}
    \caption{Qualitative examples of the output for each combination of self-attention layers modified using the proposed in-place attention reformulation. }
    \label{supp_fig:attn_layers}
    \vspace{-10pt}
\end{figure*}

\vspace{0.4em}\noindent\textbf{Resolution of attention layers:}
The denoising U-Net in the Stable Diffusion pipeline contains self-attention layers at multiple resolutions: $16\times16$, $32\times32$, and $64\times64$. 
Consequently, we can apply the proposed in-place attention reformulation to any combination of these layers.
We evaluated different combinations of these resolutions, with the results presented in \cref{supp_tab:attn_layers}. 
Additionally, we provide qualitative performance comparisons for each combination in \cref{supp_fig:attn_layers}. 
The best performance was achieved when modifying self-attention layers at all resolutions. 
This is intuitive, as it aggregates correspondences at multiple granularities, leading to a more comprehensive representation. 
In all our experiments, we use self-attention layers at all resolutions unless stated otherwise.

\section{Limitations and Future Work}\label{supp_sec:limitations}
As with other diffusion-based methods, the primary limitation of our approach lies in its high inference time. 
In this work, our focus has been on exploring the V-ICL properties of Stable Diffusion, with less emphasis on computational efficiency.  
We believe it is crucial to first establish a robust and generalizable framework, with efficiency optimizations forming an important avenue for future work.  
In particular, integrating faster diffusion techniques \citep{yin2024improved, liu2023instaflow}, which offer up to $100\times$ speedups without sacrificing output quality, could significantly reduce inference costs.

Another limitation, shared with other V-ICL methods, is sensitivity to noisy prompts.  
Since V-ICL methods rely on a small number of visual examples to infer both context and task, inaccuracies in prompt pairs can lead to degraded performance.  
While our implicitly-weighted prompt ensembling and attention temperature scaling partially mitigate this issue, further improvements in robustness to noisy or ambiguous prompts remain an open challenge.

Finally, extending our approach to the temporal domain, by adapting it to video generative models, presents a promising direction.  
Such an extension could enable training-free visual in-context learning for video-based tasks, further broadening the applicability of our framework.

Addressing these limitations could substantially enhance both the practicality and generality of visual in-context learning systems.

\section{Additional Qualitative Results} \label{supp_sec:add_qual}

\begin{figure*}
    \centering
    \includegraphics[width=0.82\linewidth]{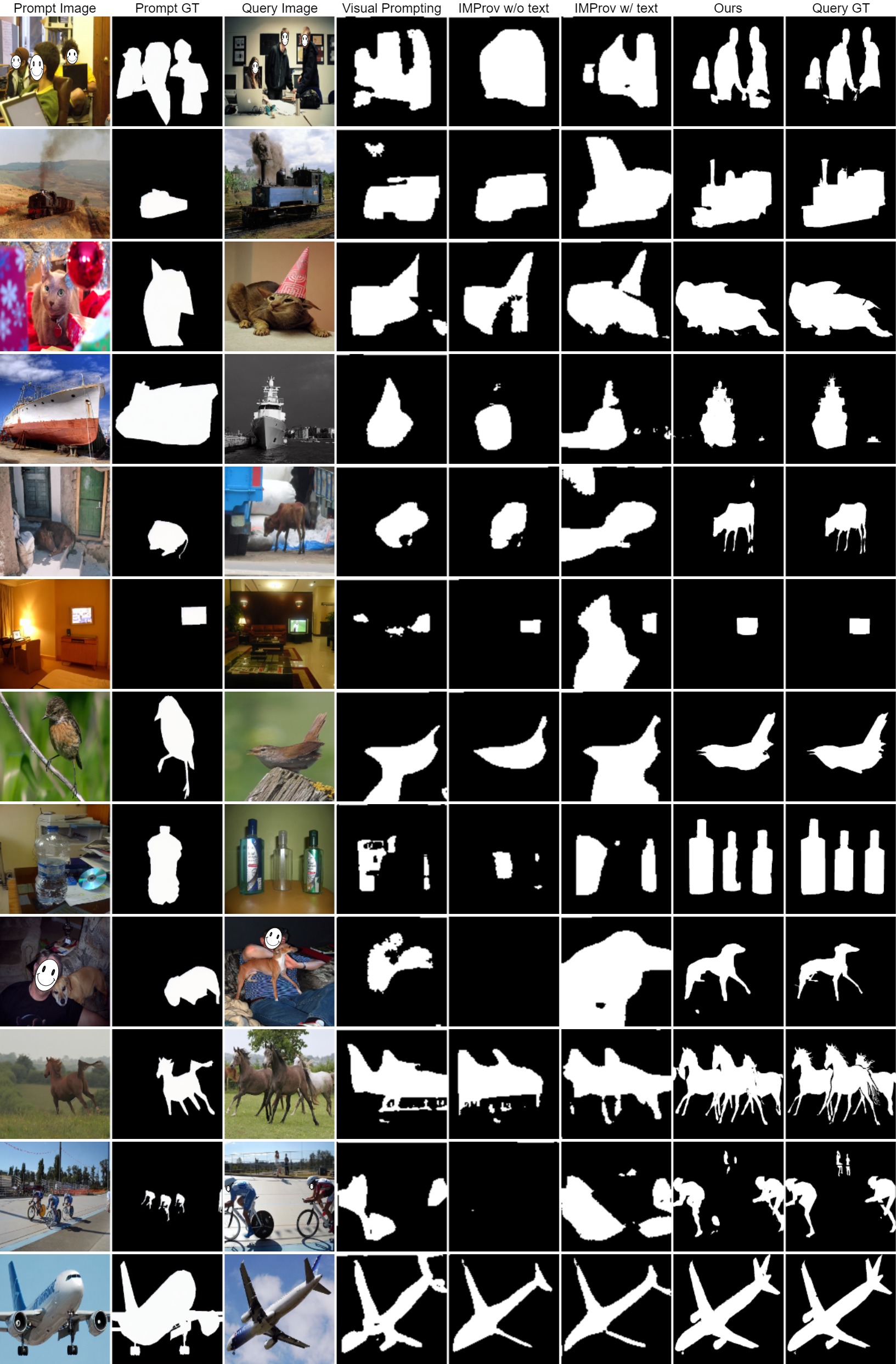}
    \caption{Qualitative examples of foreground segmentation in comparison with Visual Prompting \citep{bar2022visualprompting} and IMProv \citep{xu2023improv}. }
    \label{supp_fig:qual_fgseg}
\end{figure*}

\begin{figure*}
    \centering
    \includegraphics[width=0.82\linewidth]{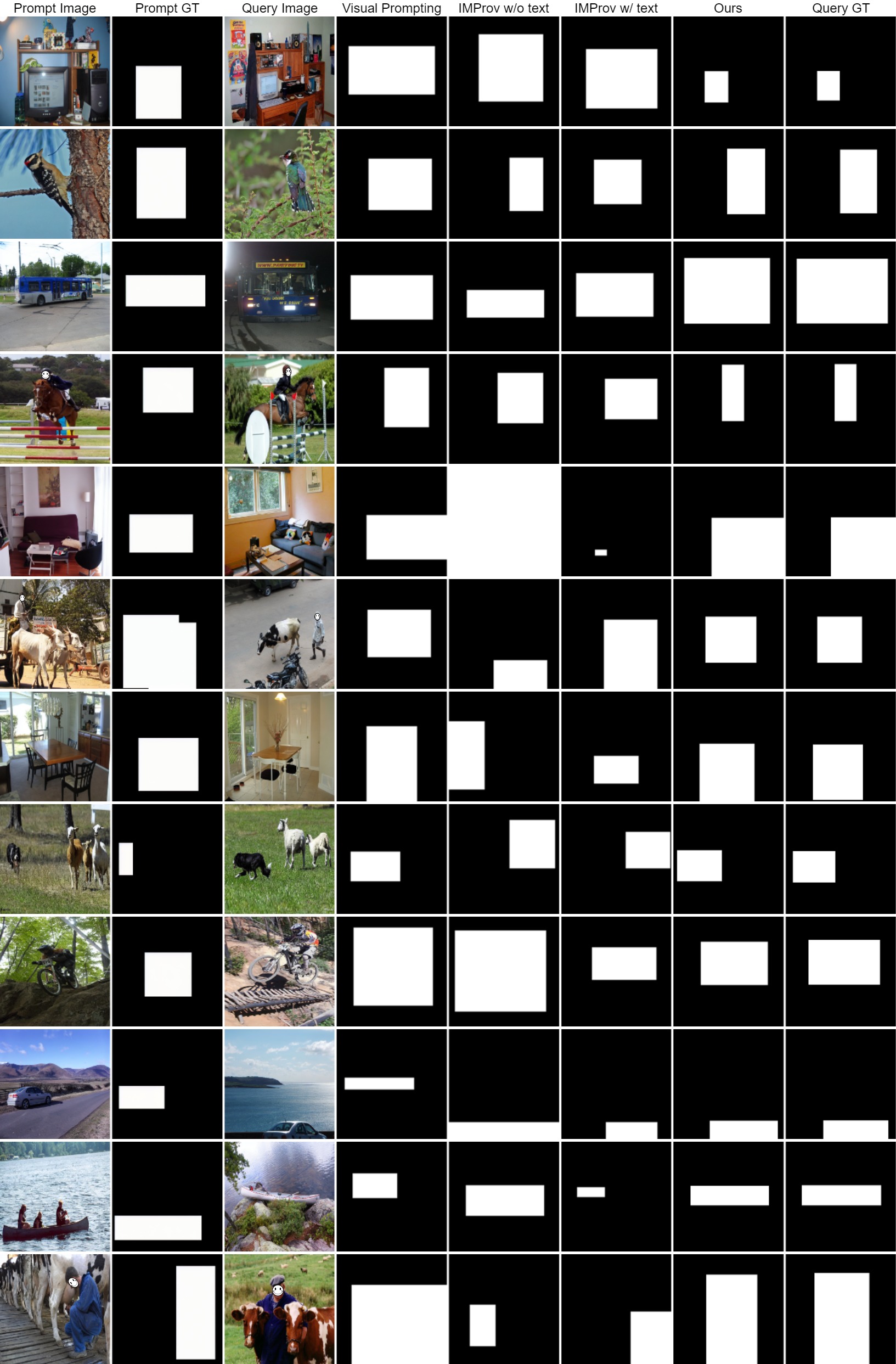}
    \caption{Qualitative examples of single object detection in comparison with Visual Prompting \citep{bar2022visualprompting} and IMProv \citep{xu2023improv}. }
    \label{supp_fig:qual_objdet}
\end{figure*}

\begin{figure*}
    \centering
    \includegraphics[width=0.82\linewidth]{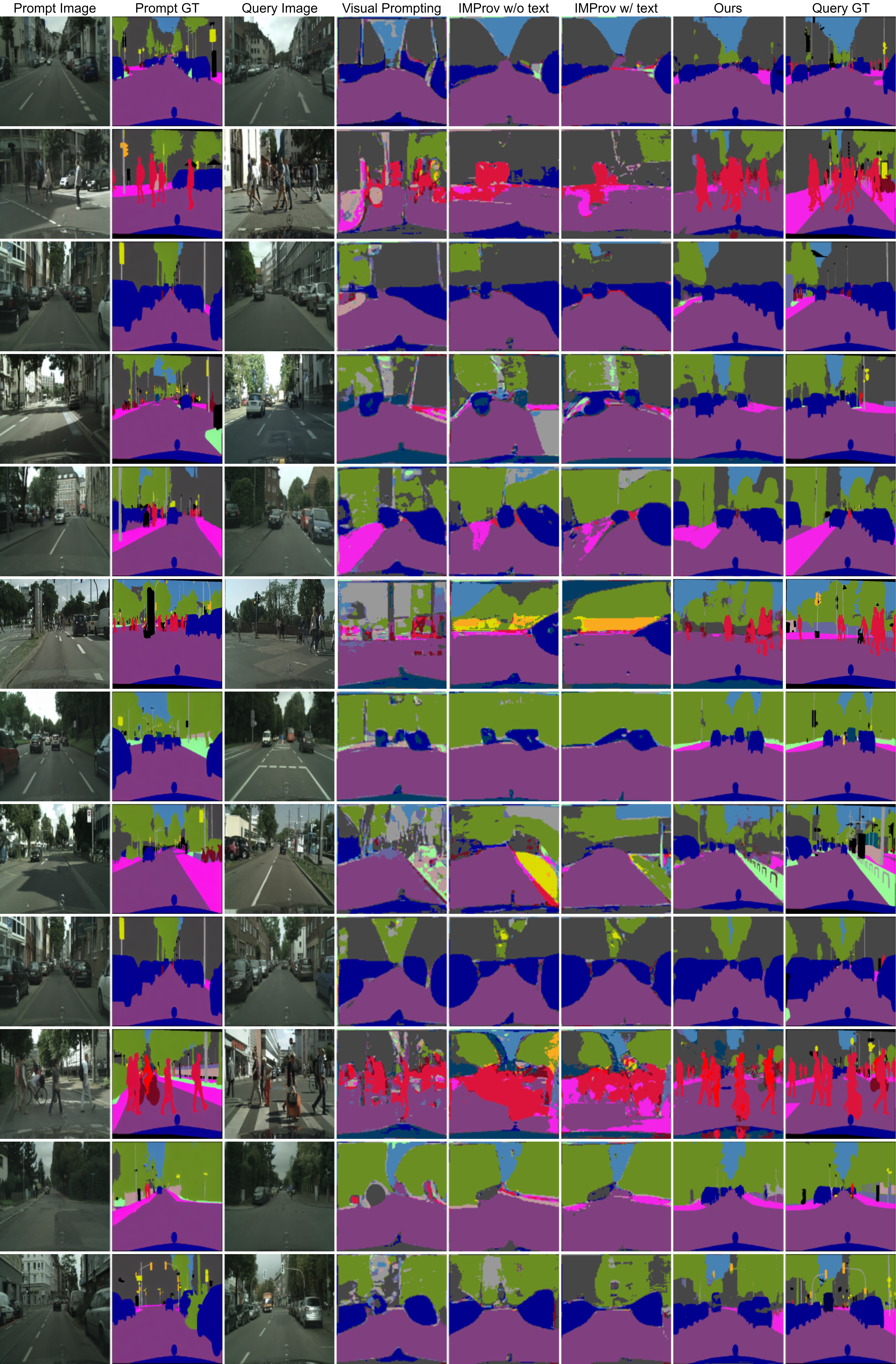}
    \caption{Qualitative examples of semantic segmentation in comparison with Visual Prompting \citep{bar2022visualprompting} and IMProv \citep{xu2023improv}. }
    \label{supp_fig:qual_semseg_cityscapes}
\end{figure*}

\begin{figure*}
    \centering
    \includegraphics[width=0.82\linewidth]{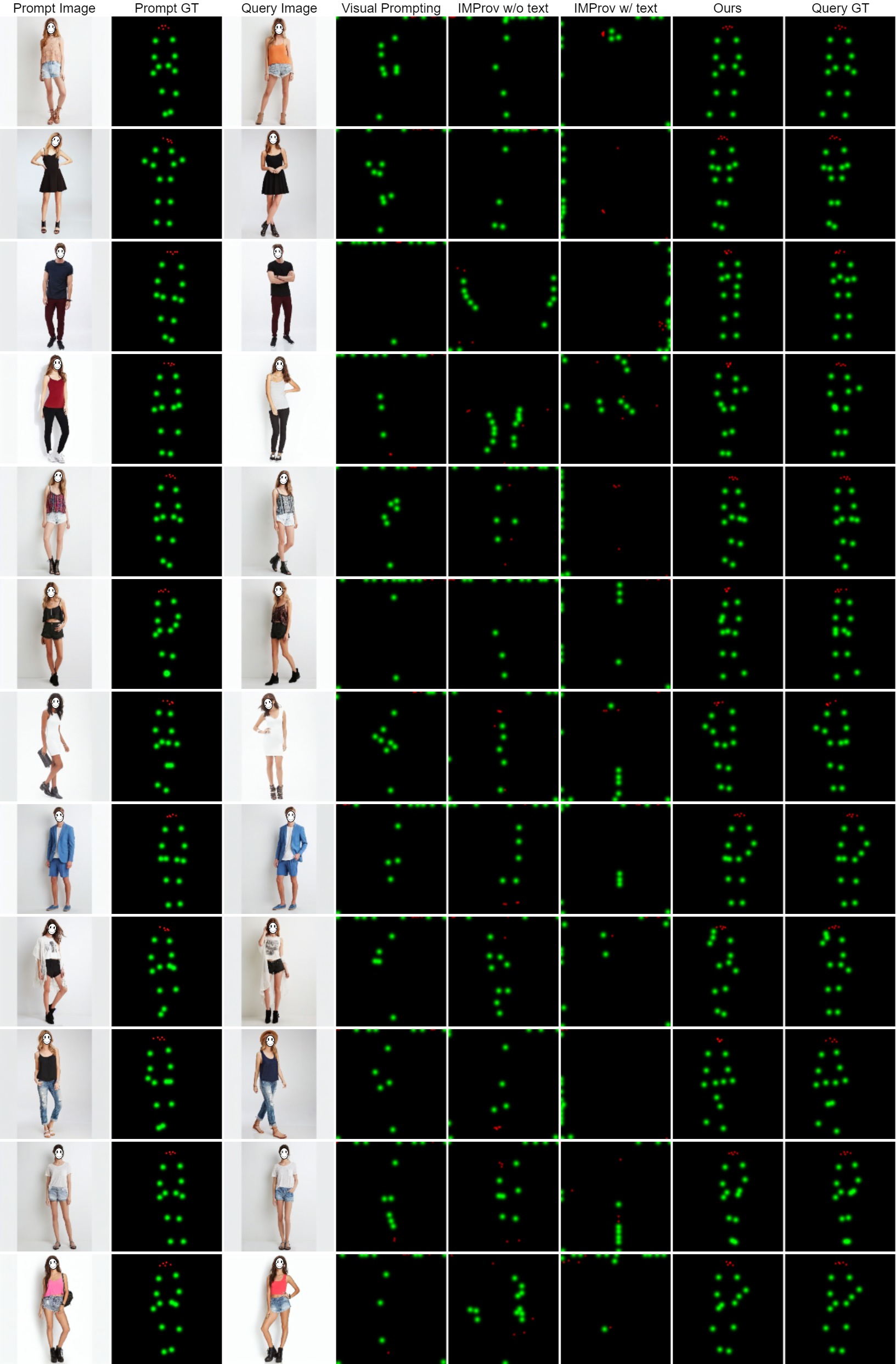}
    \caption{Qualitative examples of keypoint detection in comparison with Visual Prompting \citep{bar2022visualprompting} and IMProv \citep{xu2023improv}. }
    \label{supp_fig:qual_kpdet}
\end{figure*}

\begin{figure*}
    \centering
    \includegraphics[width=0.82\linewidth]{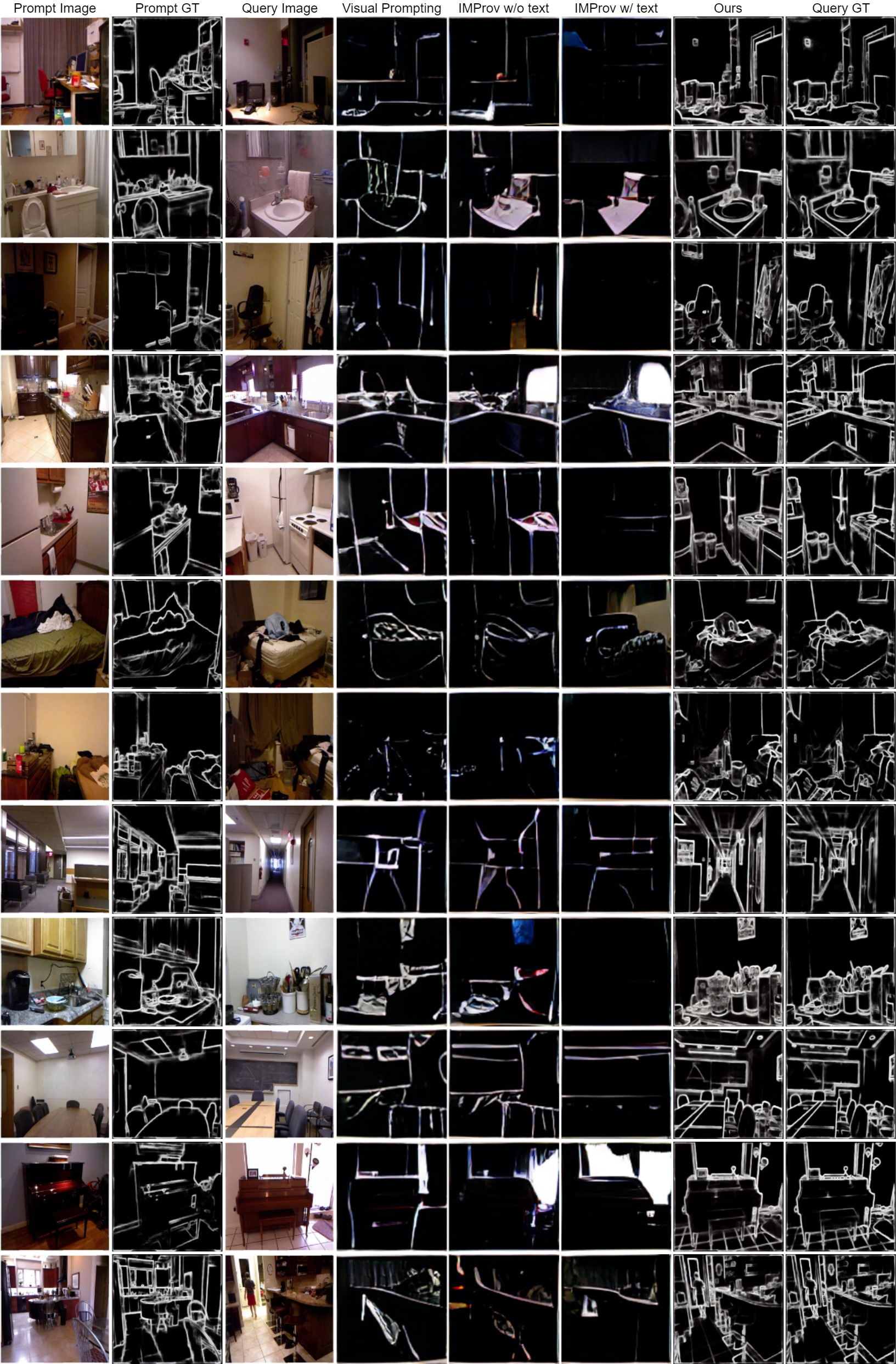}
    \caption{Qualitative examples of edge detection in comparison with Visual Prompting \citep{bar2022visualprompting} and IMProv \citep{xu2023improv}. }
    \label{supp_fig:qual_hed}
\end{figure*}

\begin{figure*}
    \centering
    \includegraphics[width=0.82\linewidth]{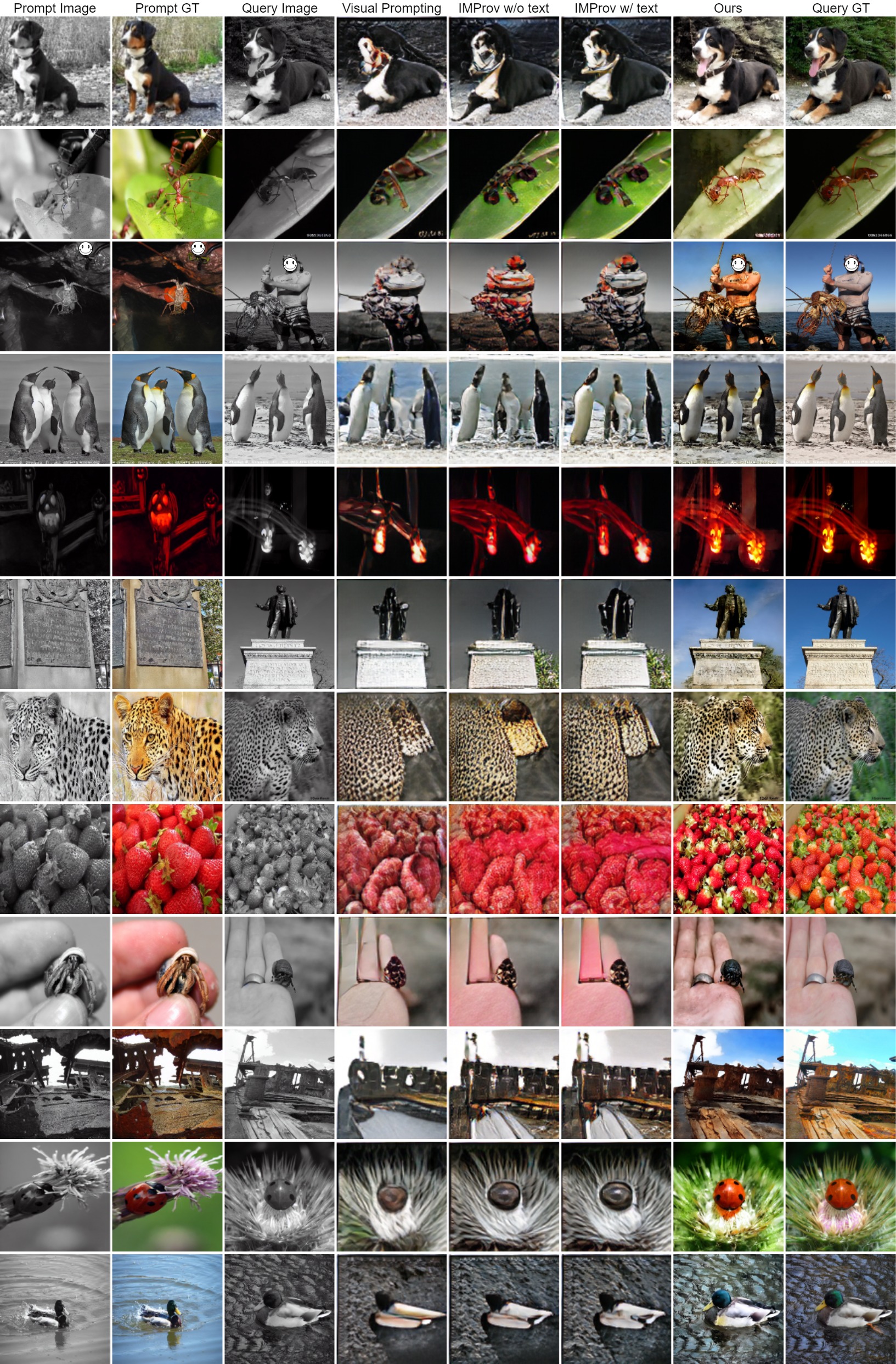}
    \caption{Qualitative examples of colorization in comparison with Visual Prompting \citep{bar2022visualprompting} and IMProv \citep{xu2023improv}. }
    \label{supp_fig:qual_clr}
\end{figure*}

We present additional qualitative examples for each task, foreground segmentation, single object detection, semantic segmentation, keypoint detection, edge detection, and colorization in \cref{supp_fig:qual_fgseg,supp_fig:qual_objdet,supp_fig:qual_semseg_cityscapes,supp_fig:qual_kpdet,supp_fig:qual_hed,supp_fig:qual_clr} respectively.

\clearpage

\end{document}